%% file: main.tex
\documentclass[sigconf]{acmart}

\usepackage{booktabs} 
\usepackage{subfig}
\usepackage{enumitem}
\usepackage{pifont}
\usepackage[table,xcdraw]{xcolor}
\usepackage{times}
\usepackage{comment}
\usepackage{color}
\usepackage{graphicx}
\usepackage{multirow}
\usepackage{breqn}
\usepackage[ruled,vlined]{algorithm2e}
\usepackage[mathscr]{euscript}

\setcopyright{rightsretained}

\acmConference[ICVGIP'25]{16th Indian Conference on Computer Vision, Graphics and Image Processing}{December 2025}{Mandi, India}
\acmYear{2025}
\copyrightyear{2025}

\acmPrice{15.00}

\begin{document}
\title{Sharing the Learned Knowledge-base to Estimate Convolutional Filter Parameters for Continual Image Restoration}

\author{Aupendu Kar}
\authornote{Work done while Aupendu Kar was at the Indian Institute of Technology Kharagpur}
\affiliation{%
  \institution{Dolby Laboratories, Inc}
  \country{Bengaluru, Karnataka, India}
}
\author{Krishnendu Ghosh}
\affiliation{%
  \institution{IIT Kharagpur}
  \country{Kharagpur, India}
}
\author{Prabir Kumar Biswas}
\affiliation{%
  \institution{IIT Kharagpur}
  \country{Kharagpur, India}
}

\renewcommand{\shortauthors}{}

\begin{abstract}
Continual learning is an emerging topic in the field of deep learning, where a model is expected to learn continuously for new upcoming tasks without forgetting previous experiences. This field has witnessed numerous advancements, but few works have been attempted in the direction of image restoration. Handling large image sizes and the divergent nature of various degradation poses a unique challenge in the restoration domain. However, existing works require heavily engineered architectural modifications for new task adaptation, resulting in significant computational overhead. Regularization-based methods are unsuitable for restoration, as different restoration challenges require different kinds of feature processing. In this direction, we propose a simple modification of the convolution layer to adapt the knowledge from previous restoration tasks without touching the main backbone architecture. Therefore, it can be seamlessly applied to any deep architecture without any structural modifications. Unlike other approaches, we demonstrate that our model can increase the number of trainable parameters without significantly increasing computational overhead or inference time. Experimental validation demonstrates that new restoration tasks can be introduced without compromising the performance of existing tasks. We also show that performance on new restoration tasks improves by adapting the knowledge from the knowledge base created by previous restoration tasks. The code is available at \href{https://github.com/aupendu/continual-restore}{https://github.com/aupendu/continual-restore}.
\end{abstract}

%
%



\maketitle

\input{samplebody-conf}

\bibliographystyle{ACM-Reference-Format}
\bibliography{mainbib}

\appendix

\end{document}

%% file: samplebody-conf.tex
\section{Introduction}
\label{sec:intro}

Due to various weather conditions and adverse natural phenomena, captured images often suffer from various types of degradation - the presence of dust and aerosols causes hazing, rain or water droplets cause poor image visibility, camera and object motion cause blurring, to name a few. These degradations diminish image quality and clarity, which in turn affect various downstream tasks, such as medical imaging ~\cite{gondara2016medical} and surveillance applications ~\cite{svoboda2016cnn}. 
\begin{figure}[!htb]
    \centering
    \includegraphics[width=0.4\textwidth]{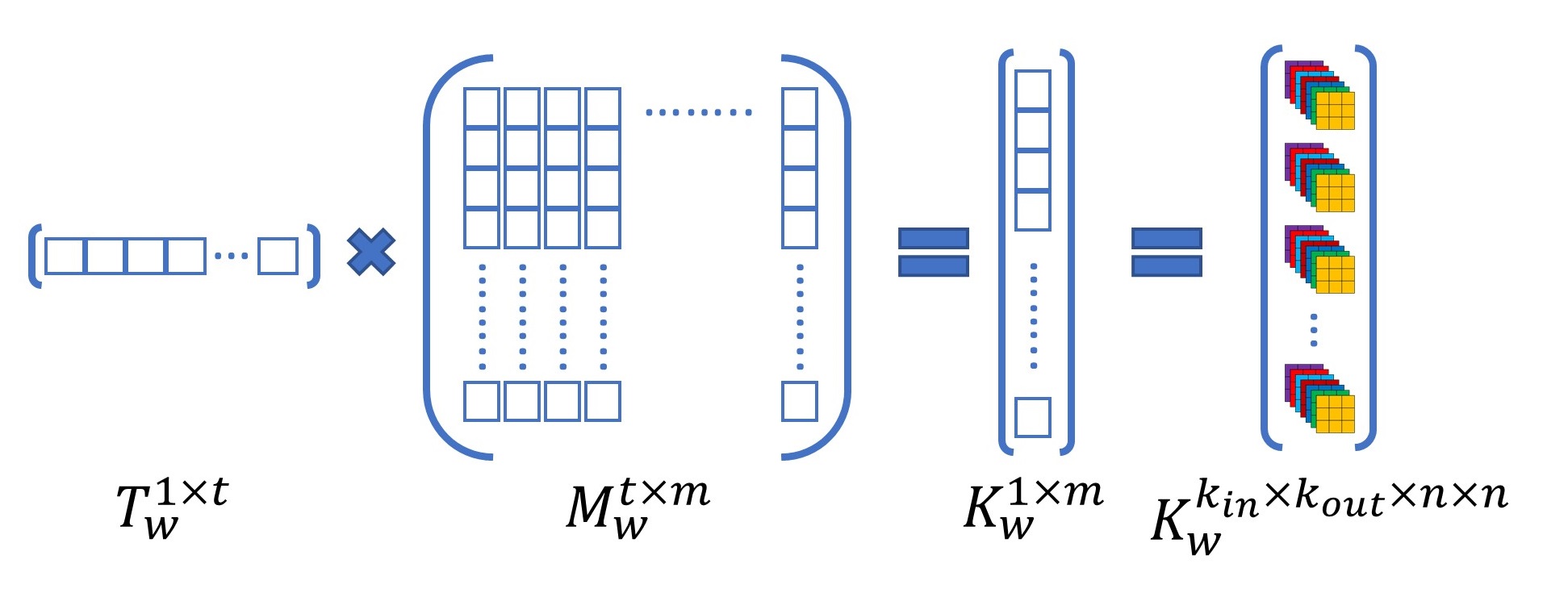}
    \vspace{-3mm}
    \caption{The proposed convolutional filter estimation block. $M_w$ is continual memory, $T_w$ is restoration task-specific weight vector, and $K_w$ is the estimated convolutional kernel.}
    \vspace{-3mm}
    \label{fig: model}
\end{figure}

Since the advent of Deep Learning, several image restoration-specific deep learning techniques have been proposed by researchers to mitigate the effects of image degradation factors, such as image dehazing~\cite{qu2019enhanced,dong2020multi}, deblurring~\cite{tao2018scale}, and deraining~\cite{Fu_2017_CVPR,deng2020detail}. Instead of designing a specific degradation model, several architectures have been proposed to handle different types of restoration~\cite{zamir2021multi,chen2021hinet}. Most of these proposed restoration methods depend entirely on the current training samples and drastically forget the learned parameters if a new restoration task is introduced, a phenomenon commonly known as catastrophic forgetting~\cite{mccloskey1989catastrophic}. This severe drawback renders a deep neural network ineffective for a previously trained task, limiting its application to the current restoration task. 

Several approaches~\cite{titsias2020functional,hu2018overcoming} have been introduced to address the problem of catastrophic forgetting. Kirkpatrick et al.~\cite{kirkpatrick2017overcoming} first proposed a parameter regularization-based algorithm where the movement of weights important to previous tasks is restricted using a quadratic constraint. Memory-aware synapses (MAS) ~\cite{MAS} also restrict the change of weights critical to previously learned tasks. It determines the relative importance of weights by computing the sensitivity of the output function with respect to each weight parameter in the network. Learning without forgetting (LwF) ~\cite{li2017learning} proposes to impose parameter regularization by using a distillation loss. A gradual pruning-based method~\cite{mallya2018packnet} is employed to compress the parameter space for a specific task and reuse the previously fixed parameters for an upcoming task. These regularization-based networks have a fixed capacity, and their performance gradually reduces as dissimilar tasks are added to the network~\cite{hung2019compacting}. These make applying them directly for image restoration tasks difficult, as the image degradation factors vary significantly in real life.

Networks with dynamic architectures, such as Progressive Neural Nets~\cite{rusu2016progressive}, aim to alleviate the problem of fixed capacity by adding new sub-networks for each new task. However, this comes with excessive computational overhead, limiting its application to edge devices with constrained computational resources.~\cite{hung2019compacting} extended the idea of PackNet~\cite{mallya2018packnet} to allow the expansion of the network by increasing the dimension of the CNN filters to accommodate new tasks. However, the simple expansion of the filter dimension causes an increase in computation in an $O(n)$ manner per pixel. In image restoration tasks, the high-resolution input images are commonly used and generally forward-propagated throughout the network architecture without any downsampling operation. Therefore, the computational burden increases significantly, and with deeper networks, the problem compounds to an even higher degree. Methods like LIRA \cite{liu2020lira} handle multiple degradations in a single image, where they utilize different task-specific expert networks for each degradation task and a common base network shared among all tasks. Due to the introduction of a new sub-network for each new task, significant computation overhead is added, and the network size increases substantially with the addition of new tasks.~\cite{Zhou_2021_CVPR} proposes lifelong learning for image restoration, focusing on a single task of deraining by allowing the network to continually learn from different rain datasets.

Distinct from these previous works, we aim to address the catastrophic forgetting problem for various image restoration tasks without incurring any significant computational burden. For this purpose, we propose a simple modification of the convolution layer, where the convolution layer in the network is factored into two parts: a task-dependent, learnable vector and a task-independent, learnable weight matrix. The task-independent weight matrix constructs a knowledge base for all restoration tasks and facilitates knowledge sharing from previous tasks by storing and reusing the earlier learned parameters for upcoming restoration tasks. In our method, the network is trained sequentially for each new restoration problem. For the first restoration task, a preassigned portion of the task-independent weight matrix is trained along with a task-dependent weight vector, the product of which generates a simple convolution filter. After completion of the training, the trained parameters from the task-independent matrix are frozen and saved. For the next task, another separate task-dependent vector is introduced, and a separate portion of the free parameters in the task-independent matrix is trained. Previously learned knowledge is reused to enhance the performance of the task at hand. This way, a very low computational overhead is incurred through a task-dependent vector for each new restoration task. The task-dependent vector also introduces a degree of freedom to the kernel generation, providing the network with the flexibility to choose or reject a previously learned filter for a new restoration task. This simple modification of the convolution layer can be easily adapted to any complex network architecture and can serve as a knowledge base for implementing continual learning-based image restoration. If the knowledge bank's parameters are exhausted, new filter kernels can be prompted by simply appending the dimension of the task-specific vector and the corresponding new dimension of the task-independent matrix. Even then, the kernel size remains the same, so no extra computational load gets added to the network.

The main contributions of this paper are as follows.

\begin{itemize}[noitemsep]
    \item We introduce a new approach to estimate the kernels of a convolutional layer, which eventually facilitates lifelong learning in image restoration tasks. To the best of our knowledge, this is the first work to deal with completely different restoration tasks in continual learning.
    \item We also demonstrate that the proposed module can be easily adapted to any other state-of-the-art network without requiring any architectural modifications.
    \item We experimentally demonstrate that the knowledge base of the proposed module can be easily expanded without incurring any significant computational burden.
    \item We experimentally validate the performance improvement in the present restoration task by using the knowledge from the previous restoration task. We also show the superiority of our proposed module as compared to similar lifelong learning approaches. 
\end{itemize}

\vspace{-3mm}

\section{Related Work}
\label{sec:literature}

\subsection{Advancement in Continual Learning}
\label{subsec:literature_continual}

The problem of catastrophic forgetting has been addressed using various methods, namely the parameter regularization method, data replay-based methods, and the dynamic network-based approach. In regularization-based methods, Kirkpatrick et al.~\cite{kirkpatrick2017overcoming} proposed a parameter regularization technique in which the weights that are relatively critical to old tasks were imposed stricter restrictions while updating for new tasks. ~\cite{Zhou_2021_CVPR} followed a similar approach to update parameters based on their importance. ~\cite{aljundi2018memory} estimates the importance by calculating the sensitivity of the output function to the change of parameters in the network. ~\cite{li2017learning,dhar2019learning} employ regularization in terms of distillation loss.~\cite{mallya2018packnet} uses iterative pruning in a trained network and fixes the previously learned critical weights using a binary mask.

Other approaches, such as \cite{shin2017continual,brahma2018subset,ostapenko2019learning,hu2018overcoming}, rely on data replay to emulate information from previous tasks. Among these, the rehearsal-based methods ~\cite{rebuffi2017icarl,lopez2017gradient,titsias2020functional} address the problem of catastrophic forgetting by remembering representative samples from the previous tasks in memory and replaying them while learning a new task. A major drawback of these methods is that previous data may not always remain available for future use. Other methods, such as \cite {shin2017continual,wu2018memory,wu2018incremental}, alleviate the problem by employing pseudo-rehearsal-based training, primarily by using generative models to generate mock samples during training for new tasks.

Dynamic network-based approaches address the forgetting problem by dedicating a portion of the network to a particular task and expanding the network as needed for new tasks. Rusu et al.~\cite{rusu2016progressive} pioneered this approach by proposing a Progressive Neural Network that prevents catastrophic forgetting by adding a sub-network for each new task and transferring previously learned features through lateral connections from the base network.~\cite{hung2019compacting} allows model expansion but maintains compactness by choosing selected learned weights by means of a learnable mask.~\cite{rosenfeld2018incremental} proposes a linear combination of existing filters to learn filters corresponding to a new task. ~\cite{li2019learn} use Neural architecture search where as ~\cite{xu2018reinforced} adopt reinforcement learning based approach for network expansion.

\subsection{Image Restoration Perspective}
\label{subsec:literature_restore}

Deep learning architectures have been used extensively in various image restoration tasks like rain-streak removal, haze removal, image denoising,
motion blur removal etc~\cite{Fu_2017_CVPR,Gandelsman_2019_CVPR,li2020zero,chang2020spatial}. Some recent works have also focused on designing a single network to perform multiple restoration tasks, rather than separate dedicated networks for domain-specific tasks ~\cite{liu2019dual,chen2021hinet,kar2021zero,zamir2021multi}. Recently, the inherent advantage of not forgetting and reusing previously acquired knowledge in continual learning (CL) has garnered interest in the restoration domain.~\cite{liu2020lira} propose a fork-join model where a new expert network that is specific to a restoration task is joined to a base pre-trained network, and a generative adversarial network is leveraged to emulate the memory replay process by generating pseudo-random samples of the previous tasks. Zhou et al.~\cite{Zhou_2021_CVPR} employ the CL mechanism for an image de-raining task by using parameter regularization based on parameters' individual importance.

\section{Methodology}
\label{sec:method}

We propose a new formulation of the convolution layer that can effectively handle multiple restoration tasks by sharing learned knowledge from previous tasks to train a new task. In this section, we discuss the proposed module, its training methodology, and the procedure for adapting previous task knowledge to new upcoming restoration tasks.
\begin{figure}[!htb]
    \centering
    \includegraphics[height=3cm]{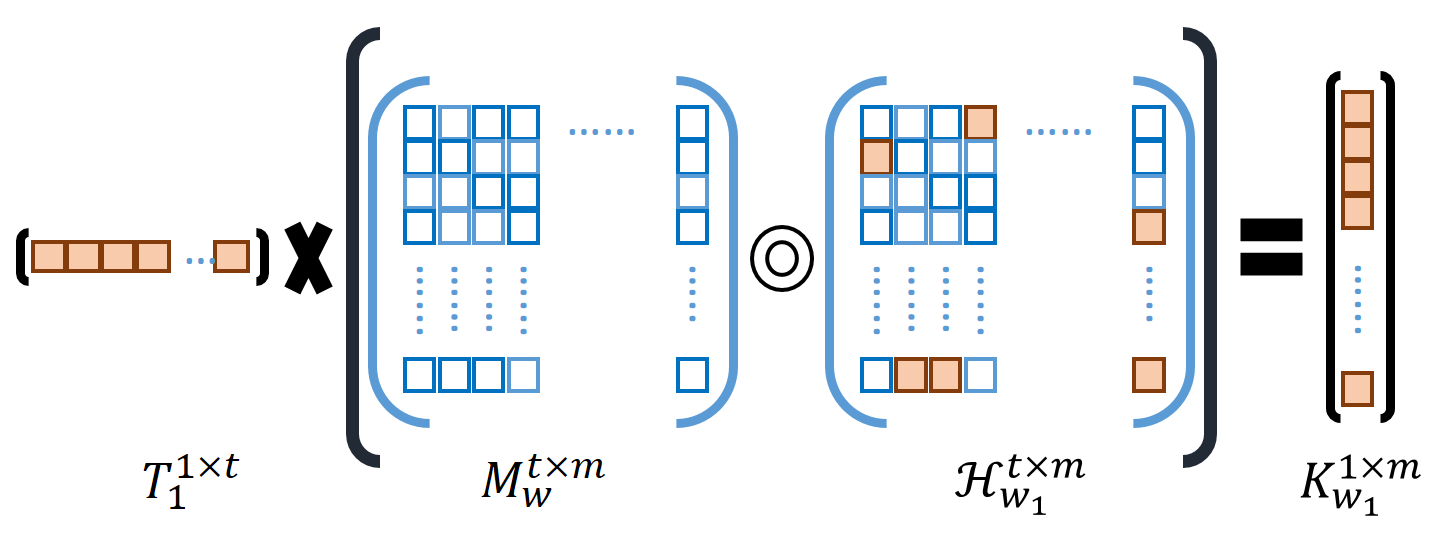}
    \vspace{-3mm}
     \caption{CMC layer during the first restoration task.}
     
  \label{fig:task_first} 
\end{figure}

\vspace{-3mm}
\subsection{Proposed Formulation of Convolution Layer}
\label{subsec:conv}

Conventional convolution layers contain learnable weights that are convolved with the input features. It can be mathematically expressed as $F_{out} = F_{in}\circledast K_{w}$, where $\circledast$ is the convolution operator, $F_{in}$ is the input feature, $F_{out}$ is the corresponding output feature, and $K_{w}$ is the kernel weights. $K_{w}$ contains trainable parameters, which are updated through a gradient back-propagation algorithm. 

Unlike the conventional method of directly determining the kernels, we estimate them indirectly by triggering a task-independent learnable weight matrix with a task-specific learnable weight vector, as shown in Figure~\ref{fig: model}. $M_w^{t\times m}$ is the task-independent weight matrix that contains the trainable weights of all the tasks. It can also be referred to as the main memory of the convolution layer, as it stores the optimized weights for various tasks. It is also expandable if the trainable parameters are exhausted. Therefore, we term it as Continual Memory in Convolution (CMC). $T_w^{1\times t}$ is the task-dependent weight vector. It is fixed for each task, and a new weight vector is introduced during the adaptation of a new upcoming task in $M_w^{t\times m}$. Here, $t$ is the length of the task-dependent weight vector. This $t$ decides the capacity of CMC. As $t$ increases, we need to add more rows in $M_w^{t\times m}$ to increase the capacity of CMC seamlessly. $m$ is the total number of parameters in a convolution kernel. The value of $m$ is mathematically expressed as $m=k_{in}.k_{out}.n.n$, where $k_{in}$ is the number of input features, $k_{out}$ is the number of output features, and $n$ is the kernel dimension. $m$ only depends on the network architecture properties. If the architecture properties are fixed, $m$ will be the same during lifelong learning. We do not need to change $m$ for any restoration task. Therefore, the main computational overhead due to convolution on input features remains unchanged for continual learning-based image restoration tasks. However, the computation may increase as we extend the dimension $t$ to expand the CMC capacity, but it is negligible compared to kernel expansion. During each task, $T_w^{1\times t}$ is matrix multiplicated with $M_w^{t\times m}$ to estimate the kernels $K_w^{1\times m}$. Both $T_w^{1\times t}$ and $M_w^{t\times m}$ contain trainable free parameters that can be trained through the gradient back-propagation algorithm. A fraction of the CMC, $M_w^{t\times m}$, is utilized in each task based on performance requirements. 
\begin{figure}[!htb]
    \centering
    \includegraphics[height=7cm]{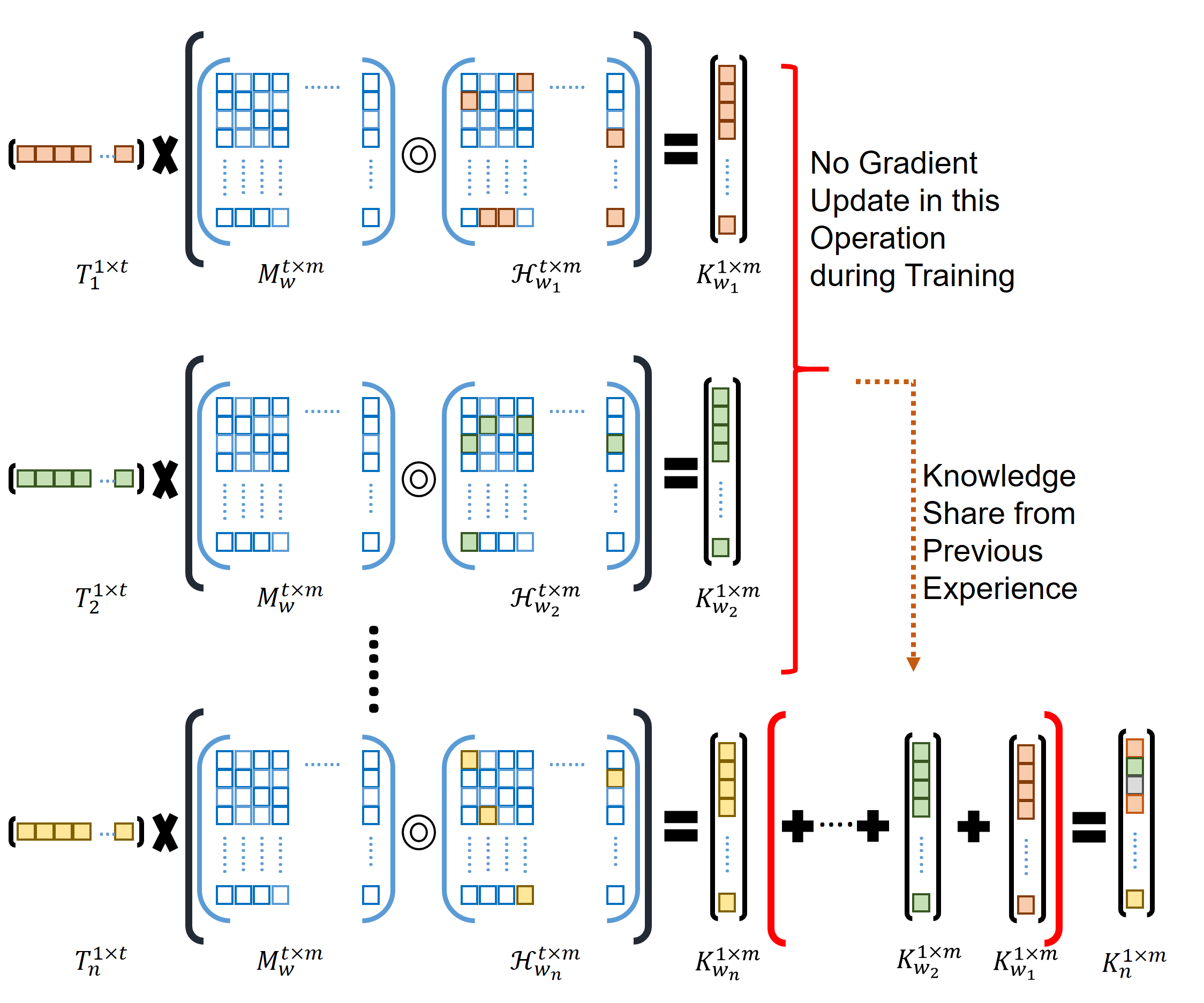}
    \vspace{-3mm}
     \caption{Operations in CMC layer for $n^{th}$ restoration task.}
  \label{fig:other_task} 
\end{figure}

\vspace{-3mm}
\subsection{Multi-task handling}
\label{subsec:multi_task}

The Continual Memory in Convolution (CMC) $M_w^{t\times m}$ is the main module whose parameters are trained in each task. In this section, the mechanism of lifelong training for restoration tasks is explained in two parts, one for the first restoration task and the other for the forthcoming restoration tasks. For the first restoration task, there is no previous knowledge to adapt. However, the forthcoming restoration tasks build upon the knowledge base established in the previous tasks. Figure~\ref{fig:task_first} shows the operations involved during adaptation of the first restoration task, and Figure~\ref{fig:other_task} shows a pictorial representation of adopting the $n^{th}$ task in the CMC module.

\subsubsection{First restoration task}

At the beginning of the first task, all the weights of the CMC module remain as free parameters. We select a random fraction of these free weights by applying a task-specific binary mask $\mathscr{H}_{w_1}^{t\times m}$ to the CMC module to train the network based on the restoration task requirements. These selected weights $M_{w_1}$ are then represented as $M_w^{t\times m}\odot\mathscr{H}_{w_1}^{t\times m}$, where $\odot$ is point-wise multiplication operator. Only these weights are expected to be updated during training for the first restoration task. In each convolution layer, restoration-specific vector $T_1$ and the selected fraction of CMC $M_{w_1}$ are updated to estimate the respective convolution kernels $K_{w_1}$, as shown in eq.\ref{eq:first}. Other weights are considered zero during this operation. After training, we get a trained $T_1$ and $M_{w_1}$.
\begin{equation}
    K_1^{1\times m} = T_1^{1\times t}.(M_w^{t\times m}\odot\mathscr{H}_{w_1}^{t\times m})
    \label{eq:first}
\end{equation}

\subsubsection{Forthcoming restoration task}

After the model is trained on the first restoration task, the forthcoming restoration tasks are trained sequentially and utilize all the trained parameters of the previous tasks, as shown in Figure~\ref{fig:other_task}. From the $2^{nd}$ task onward, the task-specific binary mask is chosen such that there is no overlap between the current mask and previously chosen masks.
\begin{equation}
   \mathscr{H}_{w_n}^{t\times m}\odot\mathscr{H}_{w_i}^{t\times m}=O , for \; all \; i = 1,...,n-1
    \label{eq:Hfree}
\end{equation}
Here $O^{t\times m}$ is a zero matrix. If $\mathscr{H}_{w}$ is the mask representing all the weights in the network, then the available free parameters in the CMC module for the $n^{th}$ task can be mathematically expressed as, 
\begin{equation}
    M_n=M_w \odot (\mathscr{H}_{w}- \mathscr{H}_{w_1}^{t\times m} \cup \mathscr{H}_{w_2}^{t\times m}\: \cup\: ...\: \cup \mathscr{H}_{w_{n-1}}^{t\times m})
    \label{eq:Mfree}
\end{equation}
The $n^{th}$ task utilizes all the filters learned from the previous tasks and learns a fraction of $M_n$ for its kernel estimation as shown in eq.\ref{eq:upcoming}.

\begin{dmath}
    K_n^{1\times m} = (T_1^{1\times t}.(M_w^{t\times m}\odot\mathscr{H}_{w_1}^{t\times m})+...+T_{n-1}^{1\times t}.(M_w^{t\times m}\odot\mathscr{H}_{w_{n-1}}^{t\times m}))+T_n^{1\times t}.(M_w^{t\times m}\odot\mathscr{H}_{w_n}^{t\times m})
    \label{eq:upcoming}
\end{dmath}
This way, the filters estimated for the $n^{th}$ task become a linear combination of previously learned filters and the newly trained kernels for the current task. All previous weights $M_{w_1},M_{w_2},...M_{w_{n-1}}$ and task specific vectors $T_1, T_2, ... T_{n-1}$ remains fixed. After training, the $n^{th}$ task occupy the fraction $M_{w_n}$ of total weight box $M_w$. 

Algorithm~\ref{algo} shows the algorithmic representation of training the proposed CMC module for the $n^{th}$ task. At the first stage, a fraction of free parameters is allocated at random for the $n^{th}$ task by using a mask $\mathscr{H}_{w_n}$. $M_w \odot \mathscr{H}_{w_n}$ represents the fraction of CMC module $M_w$, which will be tuned to learn knowledge from the $n^{th}$ task. During training, all previously learned weights are used to share the knowledge of past experiences as kernel parameters $K_{old}$ with the current task. This knowledge-sharing mechanism does not contribute to any gradient update operations. Therefore, the knowledge gained from previous experience remains unchanged. After that, $K_{old}$ is fused with task-specific kernels $K_{w_n}$ to estimate the final kernel $K_{final}$ to extract features $\mathcal{F}_{out}$ from the input features $\mathcal{F}_{in}$.

\setlength{\textfloatsep}{0pt}
\begin{algorithm}[!htb]
\SetAlgoLined
\KwIn{$T_n^{1\times t}$ = task-specific vector, $M_w^{t\times m}$ = fractionally trained CMC module}
\KwOut{Fully trained $T_n^{1\times t}$ and $M_w^{t\times m}$ with trained parameters of $n^{th}$ task}
\textit{\textbf{Allocation of Random \% of free parameters:}}\\
$\mathscr{H}_{w_1}^{t\times m}, \mathscr{H}_{w_2}^{t\times m},... \mathscr{H}_{w_{n-1}}^{t\times m}$ represents the mask of $n-1$ different tasks\;
select $\mathscr{H}_{w_n}$, where $\mathscr{H}_{w_n} \cap \mathscr{H}_{w_i}=\emptyset, \; i=1,...n-1$\;
Parameters to train $M_{w_n}=M_w \odot \mathscr{H}_{w_n}$\;
\textit{\textbf{Training on $n^{th}$ task}}\\
\ForAll{CMC layers}{
Gradient Operation Paused:\\
\For{$i\leq n-1, \; i++$}{
$K_{w_i} = T_i.(M_w \odot \mathscr{H}_{w_i})$\;
$K_{old}\; += K_i$\;
}
Gradient Operation Resumed:\\
$K_{w_n} = T_n.(M_w \odot \mathscr{H}_{w_n})$\;
$K_{final} = K_{w_n} + K_{old}$\;
$\mathcal{F}_{out} = \mathcal{F}_{in} \circledast K_{final}$
}
 \caption{Training algorithm of CMC module for $n^{th}$ task}
 \label{algo}
\end{algorithm}

\subsubsection{Extension of Parameters in a Layer}

The dimension $m$ is fixed as it is the total number of parameters that are required for a convolution. Therefore, we can increase the dimension $t$ if we exhaust all the free trainable parameters. We can also take a bigger $t$ if any layer demands more trainable parameters. For example, the input layer in a restoration task extracts key image features, and the output layer reconstructs the image from this feature domain. It may hamper performance drastically if we allocate the same percentage of weights as other layers. Unlike classification, we can not afford to lose key image features in image restoration. Our proposed modified convolution provides the flexibility to increase the parameters in those key layers without increasing computational complexity.

\section{Experimental Analysis}
\label{sec:result}

\subsection{Experimental Setting}
\label{subsec:setting}
In this section, we discuss all the experimental settings and the details of the implementation. We utilize standard datasets for various restoration tasks and employ a simple deep neural network architecture to validate our proposed idea.

\subsubsection{Restoration Task Selection}

We use four different restoration tasks for our experimental analysis. Those selected tasks are deraining, denoising, deblocking, and deblurring. These restoration tasks are chosen based on the nature of their degradation factors. Deraining is needed to alleviate the degradation caused by rain streaks. Denoising effectively reduces noise in captured images due to poor camera sensors. On the other hand, deblurring addresses the degradation caused by motion blur or poor resolution during the capture process, and deblocking mitigates blocking in an image that occurs when storing it on a disk. We explain handling these four restoration tasks throughout our paper. However, any other restoration task can be included continuously without any significant modifications.

\subsubsection{Dataset} 

We use four different datasets for the four different restoration tasks in our experiment. For deraining, we utilize the standard Rain100L dataset~\cite{yang2017deep}, which comprises $200$ training images and $100$ testing images. In the case of image denoising, we randomly add Gaussian noise with a standard deviation (std) of $50$ to the DIV2K~\cite{agustsson2017ntire} dataset, a high-resolution image dataset, for training and testing the trained model on the BSD68~\cite{roth2009fields} dataset. In both deblocking and deblurring, we use the DIV2K dataset and degrade the image with random JPEG artifacts and blurring, respectively, during training. We consider the quality range $[10, 70]$ for introducing the JPEG artifact. For blurring, we take account of the Gaussian blur and take $15 \times 15$ blur kernel with a random standard deviation in the range $[0.2, 3]$. However, during testing, we only considered the Gaussian blur kernel with a standard deviation of $2.5$ for deblurring and the insertion of JPEG artifacts using a quality factor of $20$ for deblocking. In both deblocking and deblurring, we utilize the DIV2K validation dataset for testing purposes.

\subsubsection{Model Architecture}

We use a simple consecutive residual block-based network architecture~\cite{he2016deep} for our experimental purposes. There are $6$ residual blocks in our network, excluding the input and output convolution blocks. Each residual block consists of $64$ input and output channels. The convolution blocks inside each residual block use $3\times 3$ convolution with stride $1$. In our experiment, we replace the conventional convolution blocks with our proposed modified block. However, all the kernel parameters remain the same. Therefore, it can be seamlessly integrated into any deep architecture without requiring any architectural modifications.

\subsubsection{Implementation details}

We use the same experimental setup for all the experiments. The model is trained for $125$ epochs, and each epoch consists of $1,000$ batch updates. There are $16$ image patches of size $128 \times 128$ in each batch. All images are normalized to the range $[0, 1]$ during both training and testing. The mean-squared error (MSE) is used as a loss function for gradient back-propagation. Adam optimizer ~\cite{kingma2014adam} with learning rate $10^{-4}$ is used for updating the weights, and the learning rate is halved after every $25$ epochs. We use the Peak signal-to-noise ratio (PSNR) metric throughout the paper for performance analysis.
\begin{table}[!htb]
\centering
\resizebox{0.45\textwidth}{!}{%
\begin{tabular}{|c|c|c|c|c|c|}
\hline
\begin{tabular}[c]{@{}c@{}}\%\\ Params\end{tabular} & \begin{tabular}[c]{@{}c@{}}Knowledge\\ Sharing\end{tabular} & Derain & Denoise & Deblocking & Deblur \\ \hline
\multirow{2}{*}{20}   & \ding{55}  & 33.50 & 27.65 & 30.99 & 29.64 \\ \cline{2-6} 
                      & \ding{51} & 33.50 & \textbf{27.68} & \textbf{31.11} & \textbf{29.74} \\ \hline
\multirow{2}{*}{10}   & \ding{55}  & 32.14 & 27.43 & 30.81 & 29.32 \\ \cline{2-6} 
                      & \ding{51} & 32.14 & \textbf{27.53} & \textbf{30.96} & \textbf{29.64} \\ \hline
\multirow{2}{*}{5}    & \ding{55}  & 30.36 & 27.08 & 30.55 & 28.97 \\ \cline{2-6} 
                      & \ding{51} & 30.36 & \textbf{27.23} & \textbf{30.78} & \textbf{29.33} \\ \hline
\multirow{2}{*}{2.5}  & \ding{55}  & 29.71 & 26.58 & 30.29 & 28.35 \\ \cline{2-6} 
                      & \ding{51} & 29.71 & \textbf{27.01} & \textbf{30.64} & \textbf{29.02} \\ \hline
\multirow{2}{*}{1.25} & \ding{55}  & 29.19 & 26.10 & 30.21 & 27.92 \\ \cline{2-6} 
                      & \ding{51} & 29.19 & \textbf{26.49} & \textbf{30.42} & \textbf{28.52} \\ \hline
\end{tabular}%
}
\caption{Performance of continual task adaptation on PSNR metric. \ding{51} means knowledge of previous tasks is adapted during training. Derain is the first task. Denoise, deblocking, and deblur are the next tasks on which the model is trained, following that sequence.}
\vspace{-6mm}
\label{tab:continual}
\end{table}

\subsection{Experiments on Continual Task Adaptation}
\label{subsec:perform}

Table~\ref{tab:continual} shows the quantitative analysis of lifelong restoration task learning with knowledge sharing. In the first experiment, we allocate parameters for different tasks and train on restoration datasets without sharing knowledge from other tasks. In the second experiment, the knowledge of past restoration tasks is shared with the current tasks. This way, lifelong learning persists. There is no performance difference in single-image deraining as it is the first task. Denoising, deblocking, and deblurring are the next consecutive tasks. We observe from the table that performance on these three tasks consistently yields better results when knowledge is shared. We can also observe that knowledge sharing performs significantly better as the percentage of parameter allocation decreases. This happens because decreasing the allocated parameters hinders the learning process, and the model can not acquire sufficient knowledge for that particular task. Therefore, similar knowledge of previous restorations becomes more helpful in learning the current restoration task. 

\begin{figure}[!htb]
    \centering
    \subfloat[\centering $5\%$ model parameters used]{{\includegraphics[width=0.23\textwidth]{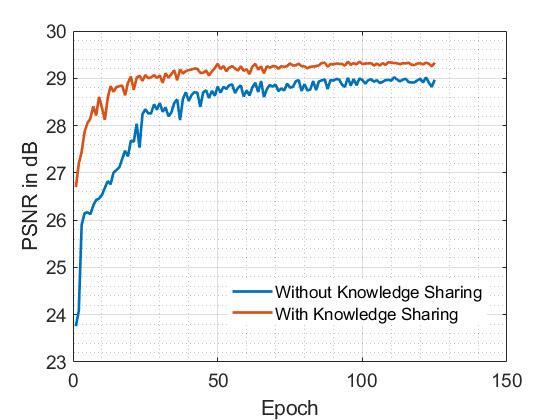} }}
    \subfloat[\centering $1.25\%$ model parameters used]{{\includegraphics[width=0.23\textwidth]{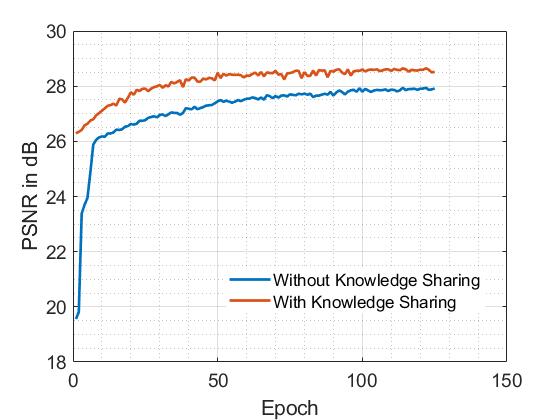} }}%
    \vspace{-3mm}
    \caption{PSNR in dB vs Each epoch of training}%
    \label{fig: continual}%
    \vspace{-3mm}
\end{figure}

Figure~\ref{fig: continual} shows how the performance metric PSNR changes in each training epoch. We chose the final restoration task, deblurring, for this analysis purpose. Figure~\ref{fig: continual}(a) shows performance analysis with $5\%$ of model parameters, and Figure~\ref{fig: continual}(b) shows performance analysis with $1.25\%$ of model parameters. We can clearly see the improvement by applying the knowledge gained from deraining, denoising, and deblocking, as the PSNR in the first epoch already yields an initial difference of approximately $3.5$ dB for $5\%$ parameters and approximately $6$ dB for $1.25\%$ parameters. Therefore, we can say that previous task knowledge gives better performance and results in faster convergence.

\begin{table}[!htb]
\centering
\resizebox{0.45\textwidth}{!}{%
\begin{tabular}{|l|c|c|c|c|}
\hline
Methods  & Derain & Denoise & Deblocking & Deblur \\ \hline
Together & 25.44  & 15.66   & 29.26      & 29.80  \\ \hline
Deform   & 30.90  & 27.26   & 30.67      & 29.29  \\ \hline
Pruning~\cite{mallya2018packnet}  & 29.56  & 27.45   & 31.00      & 29.50  \\ \hline
MAS~\cite{MAS}  & 23.80  & 18.99   & 28.39      & 25.98  \\ \hline
CMC-5    & 32.14  & 27.53   & 30.96      & 29.64  \\ \hline
\end{tabular}
}
\caption{Quantitative analysis of our proposed CMC module with other continual learning mechanisms.}
\vspace{-3mm}
\label{tab:pruning}
\end{table}

\subsection{Comparative Analysis}
\label{subsec:comp}

Most of the continual learning based frameworks are specifically designed for classification tasks. Therefore, it is not feasible to apply those to restoration models. However, for comparative analysis, we evaluate different baseline models and popular pruning-based continual learning methods. Table~\ref{tab:pruning} shows the quantitative evaluation of our proposed Continual Memory in Convolution (CMC) module with different baseline models. The pruning-based methods directly prune the filters of the convolution layers and use those pruned weights for upcoming tasks~\cite{mallya2018packnet}. The `Deform' baseline utilizes a deformable convolution-based architecture to reduce the model's parameters. This baseline aims to compare the advantage of a model with shared knowledge with that of smaller, distinct models for various tasks. In the deform baseline, separate models are used for different tasks. The `Together' baseline model incorporates all tasks into a single model. This baseline uses the whole ResNet architecture to train the model for all four tasks. The `Deform' baseline utilizes around $13\%$ more model parameters compared to a plain convolution-based architecture. The `Pruning' baseline uses $12.5\%$ of model parameters, and our proposed CMC-5 takes $10\%$ of the overall model parameters. If we consider the total number of parameters, our model has more parameters as compared to pruning-based methods. However, the kernel parameters and throughput speed remain the same. MAS~\cite{MAS} is a regularization-based continual learning method that can be easily applied to restoration tasks. For a fair comparison, we use the same residual block-based architecture to perform the experiments using MAS. MAS failed to maintain its performance in past restoration tasks. After training on all four restoration tasks sequentially, the PSNR drops significantly from $34.36$ dB to $23.80$ dB for deraining, from $25.42$ dB to $18.99$ dB for denoising, and from $30.11$ dB to $28.39$ dB for deblocking. 

\subsection{Additional Analysis}
\label{subsec:other}

\begin{figure*}[!htb]
   \centering
   \subfloat[\centering Grouth-Truth]
   {
   \includegraphics[width=0.23\textwidth]{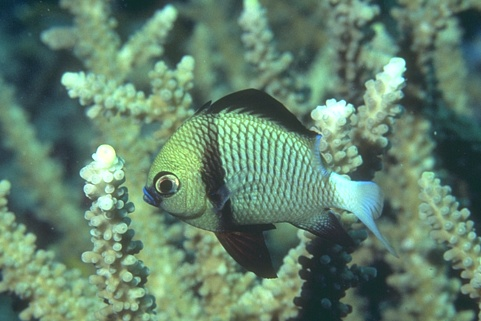}
   }%
   \subfloat[\centering Rainy Image]{\includegraphics[width=0.23\textwidth]{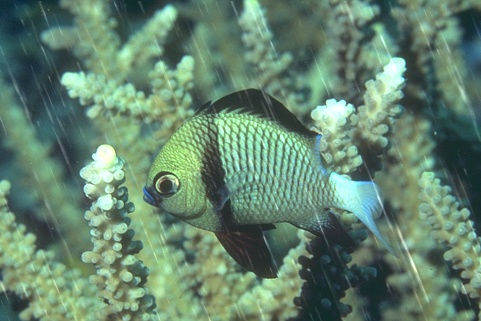} }%
   \subfloat[\centering Deraining (Image-1)]{\includegraphics[width=0.23\textwidth]{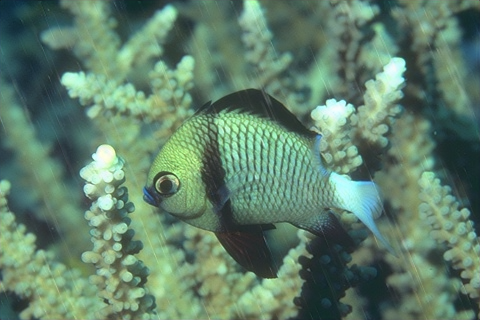} \label{subfig:derain1} }%
   \subfloat[\centering Deraining (Image-2)]{\includegraphics[width=0.23\textwidth]{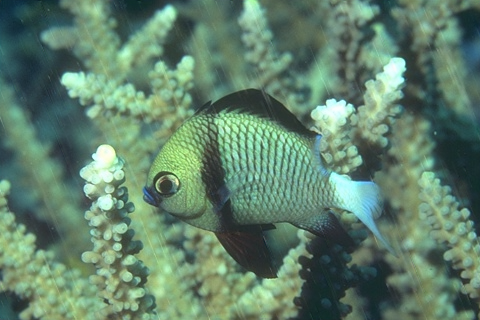} \label{subfig:derain2} }%
   \vspace{-3mm}
   
   \subfloat[\centering Grouth-Truth]{\includegraphics[width=0.23\textwidth]{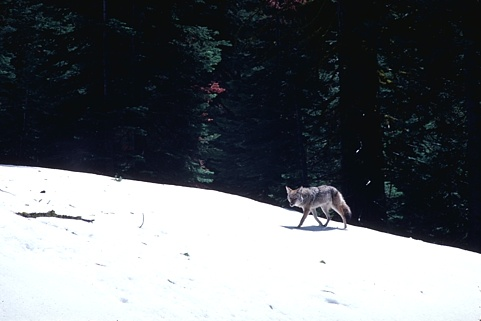} }%
   \subfloat[\centering Noisy Image]{\includegraphics[width=0.23\textwidth]{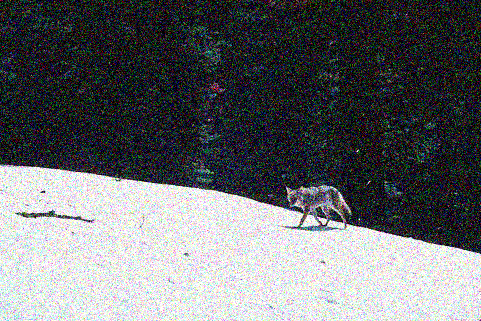} }%
   \subfloat[\centering Denoising (Image-1)]{\includegraphics[width=0.23\textwidth]{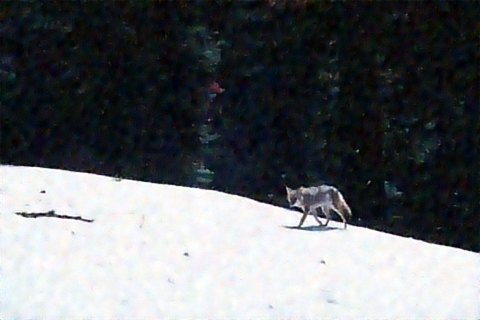} \label{subfig:denoise1} }%
   \subfloat[\centering Denoising (Image-2)]{\includegraphics[width=0.23\textwidth]{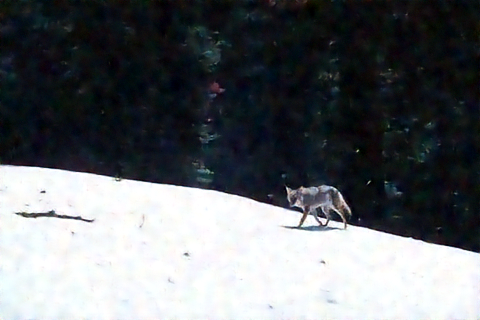} \label{subfig:denoise2} }%
   \vspace{-3mm}
   
   \subfloat[\centering Grouth-Truth]{\includegraphics[width=0.23\textwidth]{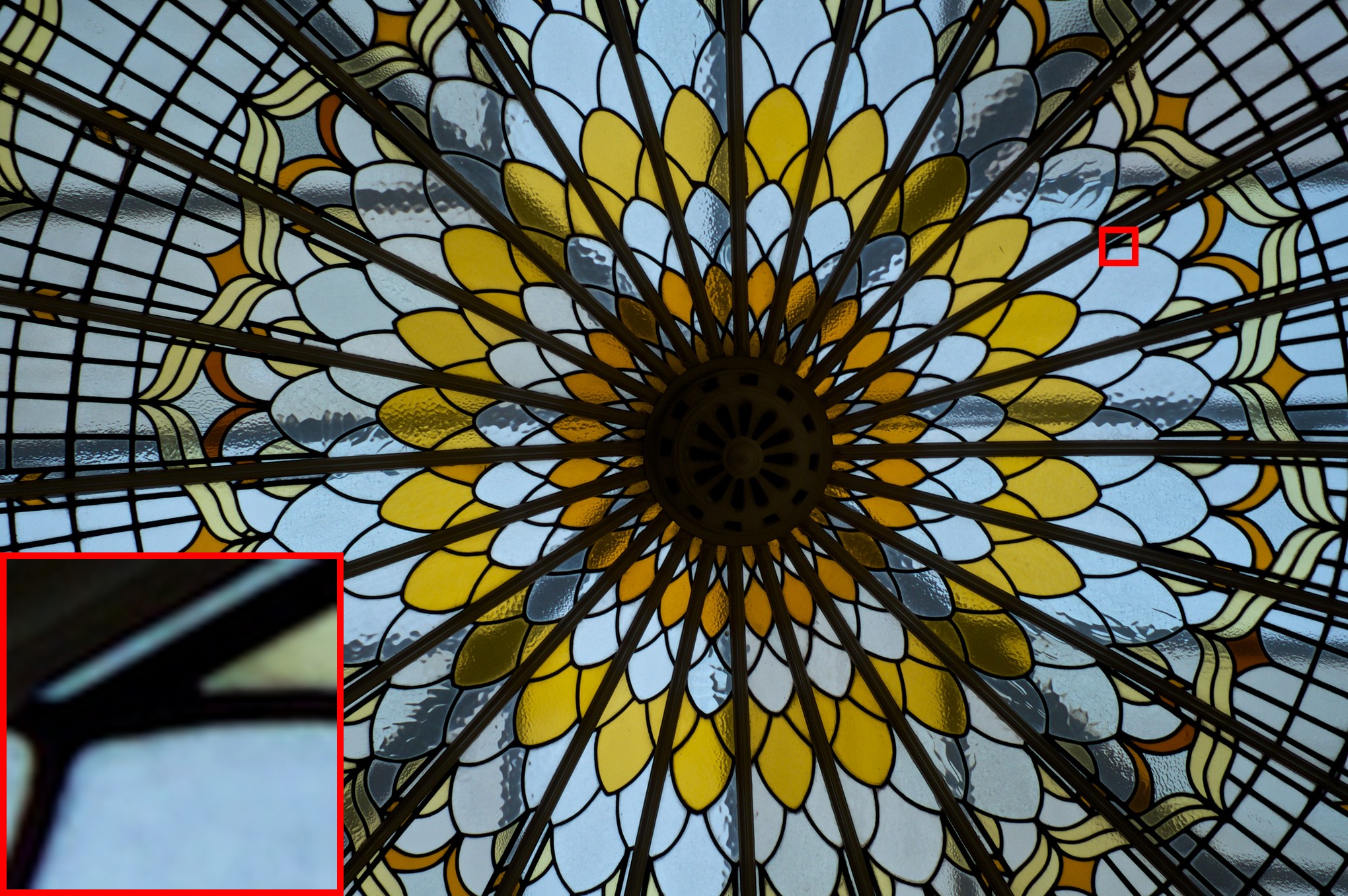} }%
   \subfloat[\centering Blury Image]{\includegraphics[width=0.23\textwidth]{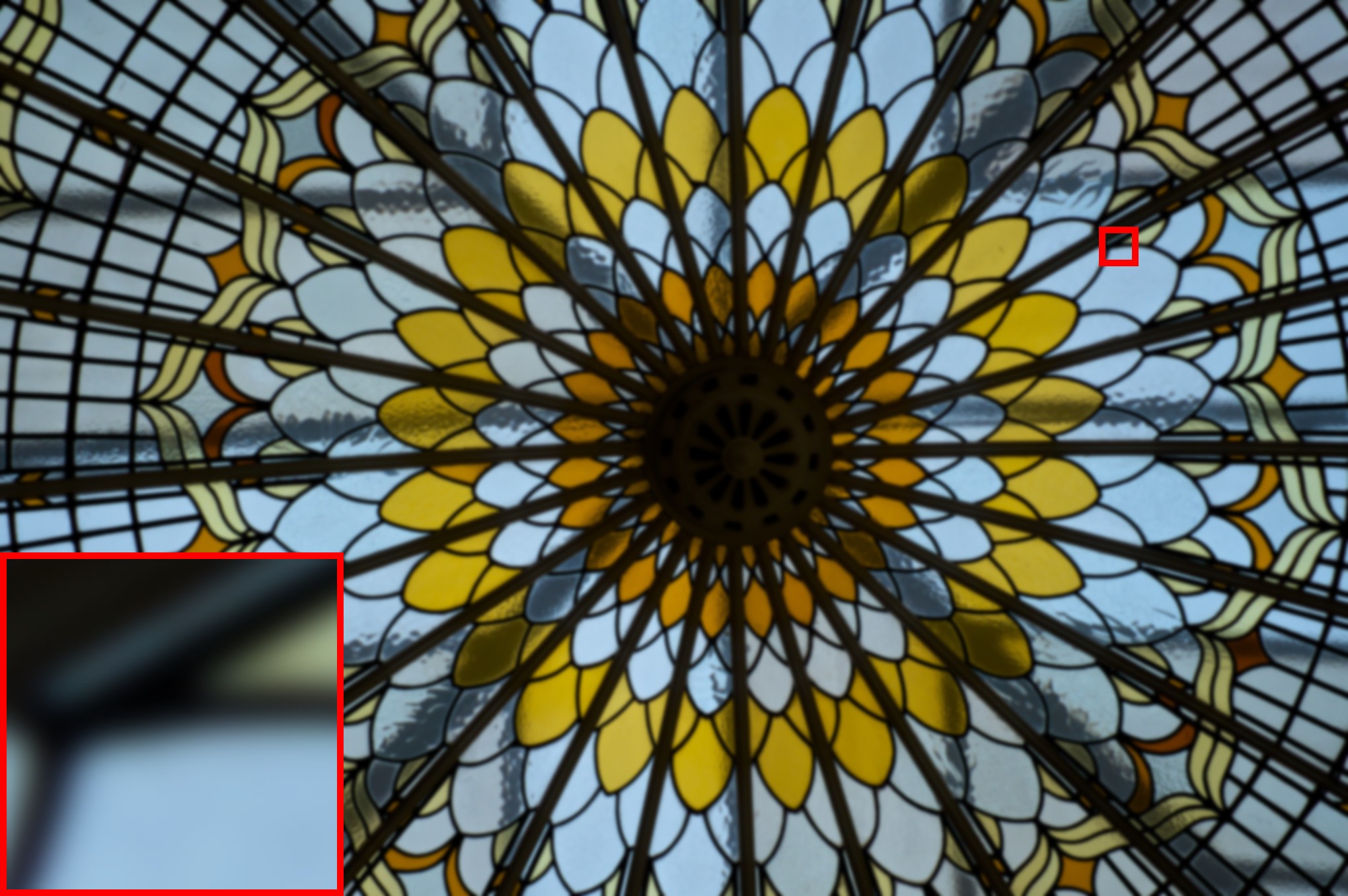} }%
   \subfloat[\centering Deblurring (Image-1)]{\includegraphics[width=0.23\textwidth]{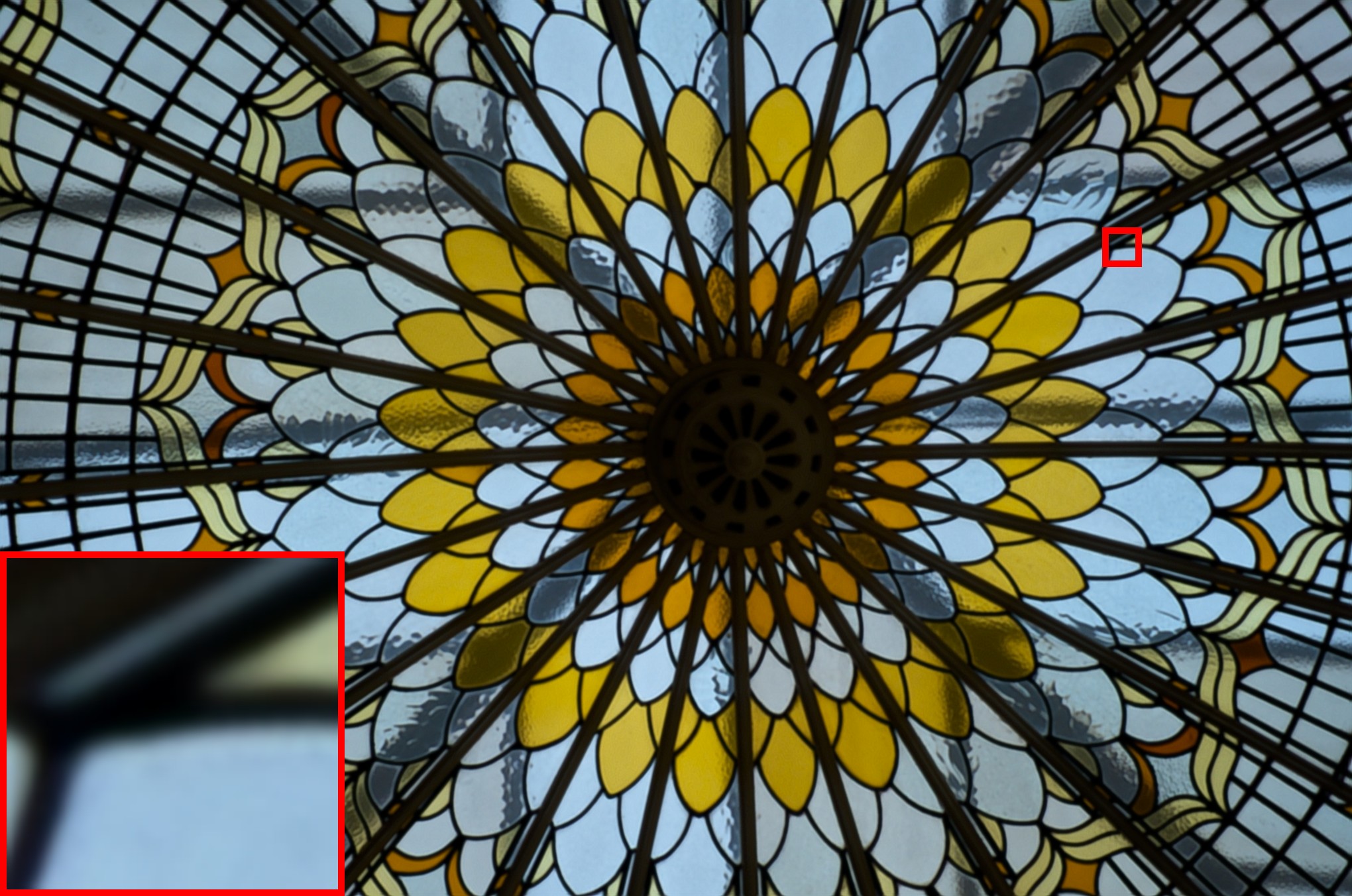} \label{subfig:deblur1} }%
   \subfloat[\centering Deblurring (Image-2)]{\includegraphics[width=0.23\textwidth]{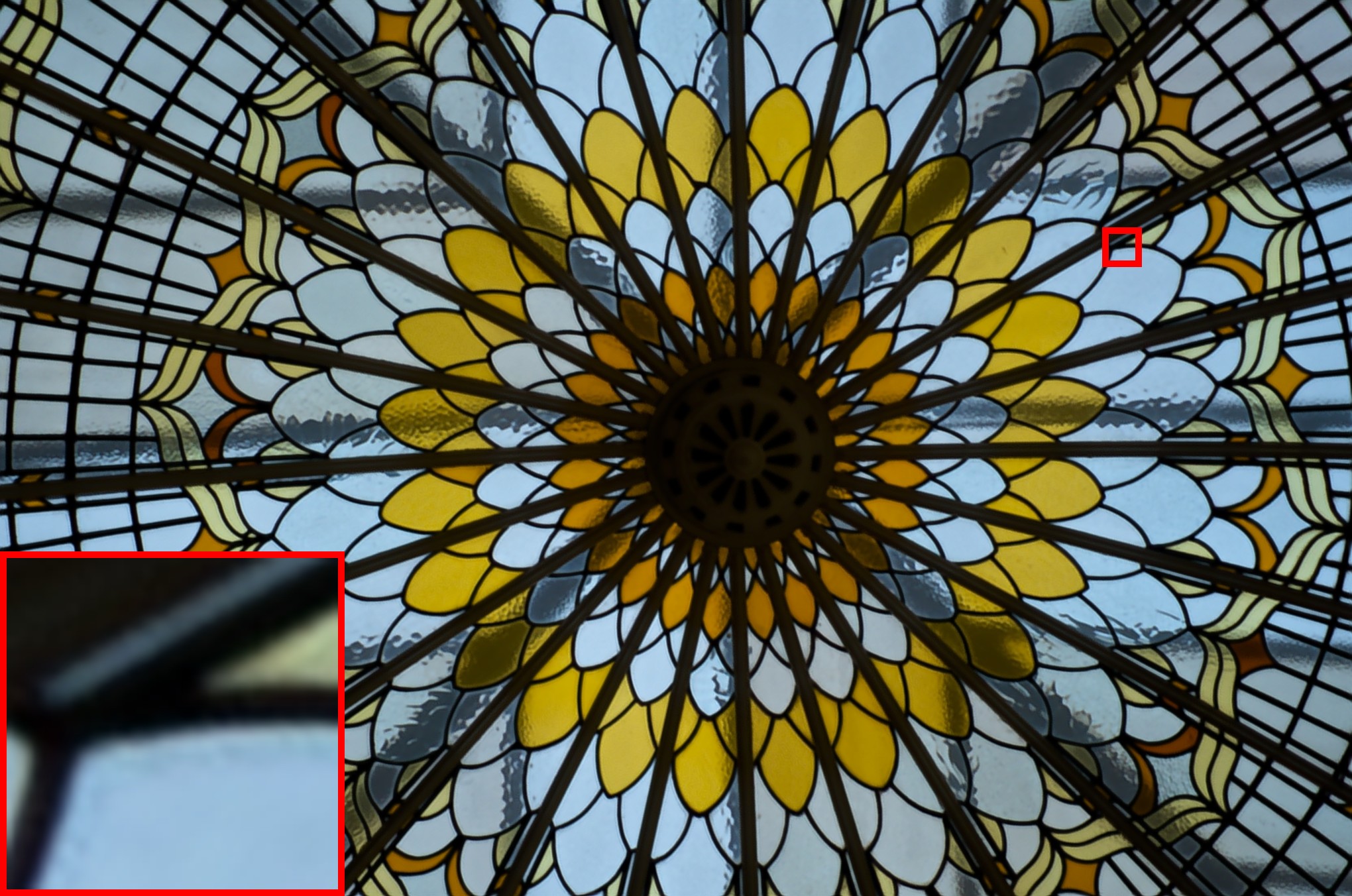} \label{subfig:deblur2} }%
   \vspace{-3mm}
   
   \subfloat[\centering Grouth-Truth]{\includegraphics[width=0.23\textwidth]{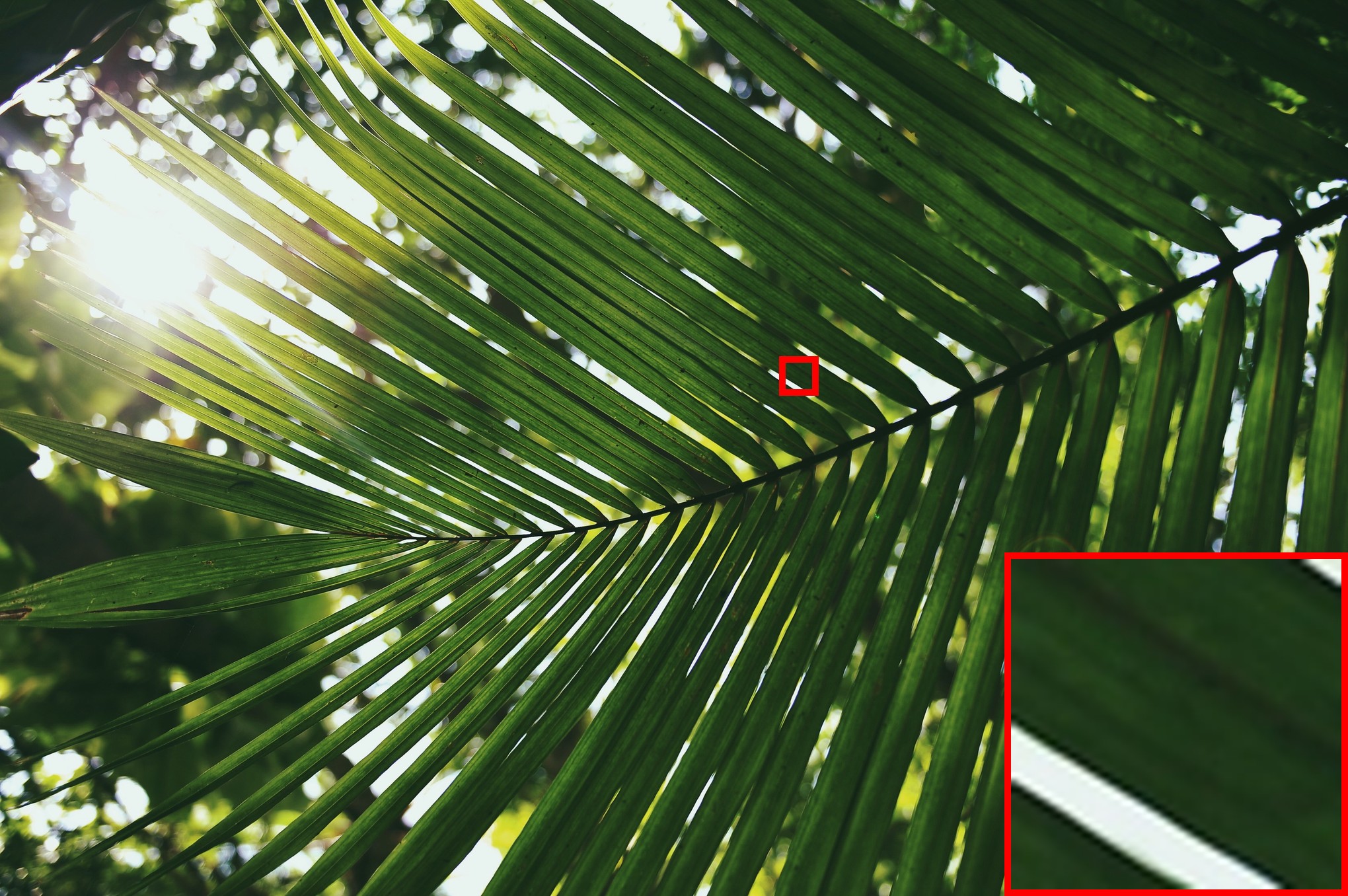} }%
   \subfloat[\centering JPEG Artifact Image]{\includegraphics[width=0.23\textwidth]{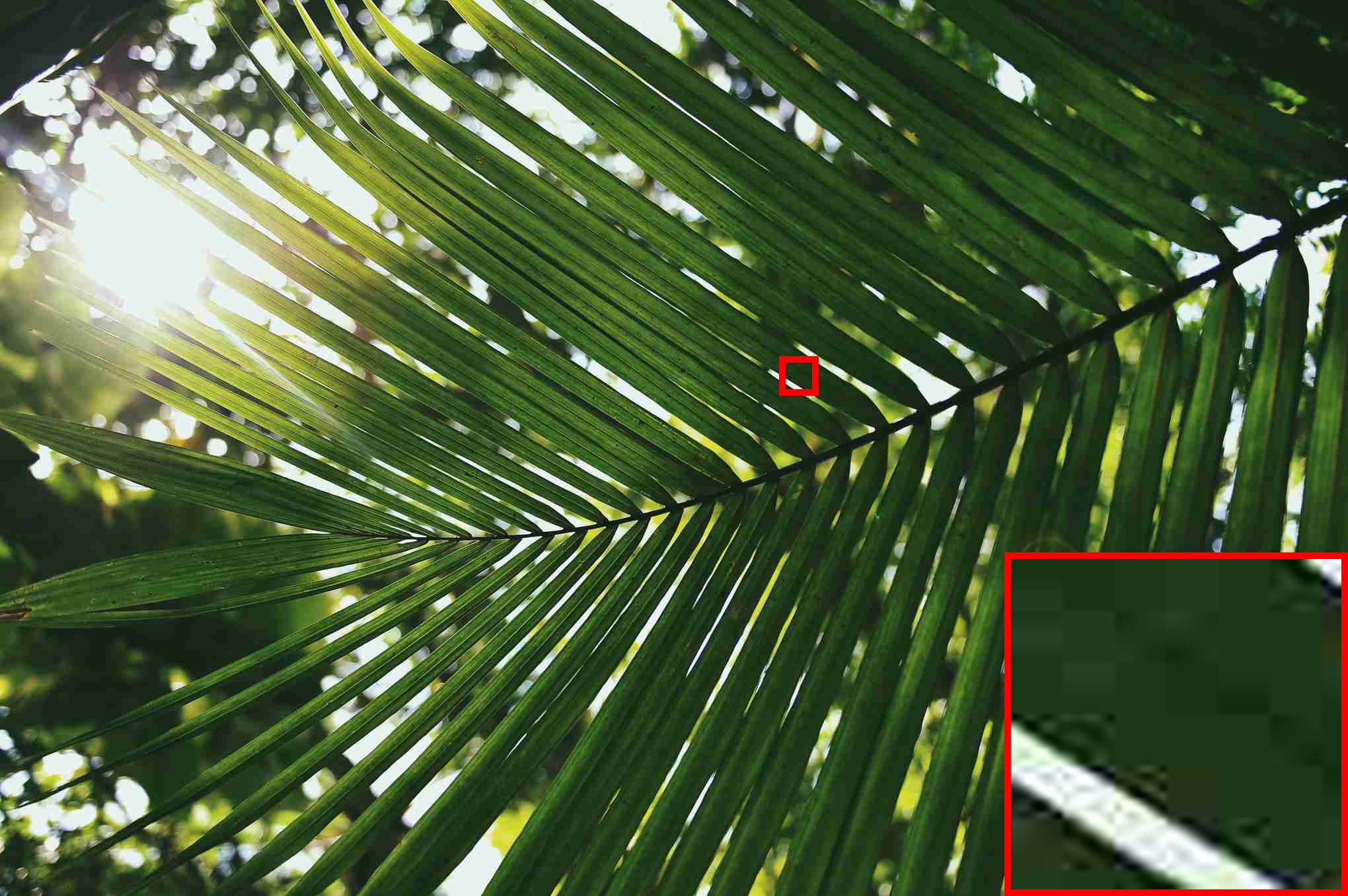} }%
   \subfloat[\centering Deblocking (Image-1)]{\includegraphics[width=0.23\textwidth]{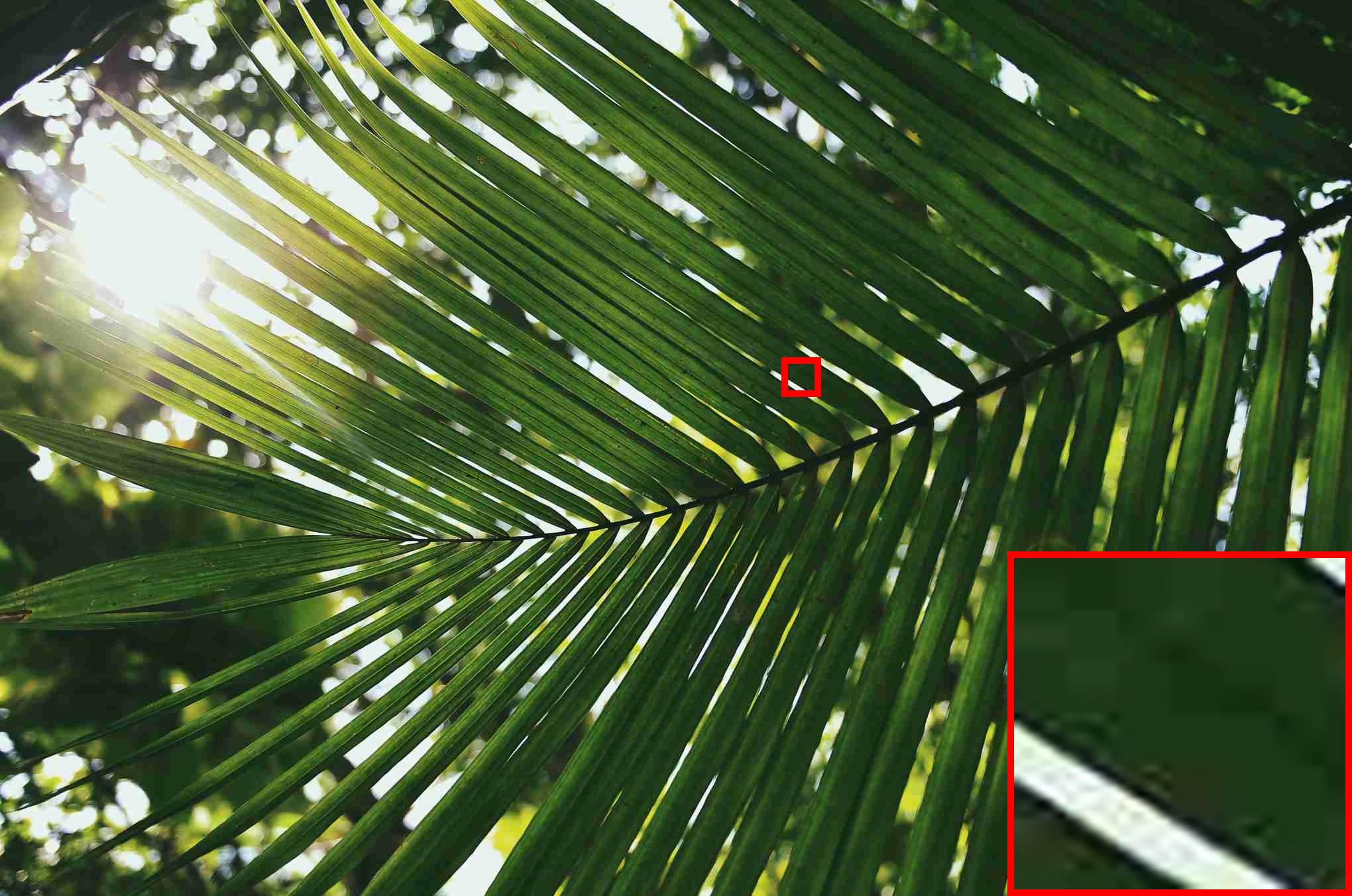} \label{subfig:deblock1} }%
   \subfloat[\centering Deblocking (Image-2)]{\includegraphics[width=0.23\textwidth]{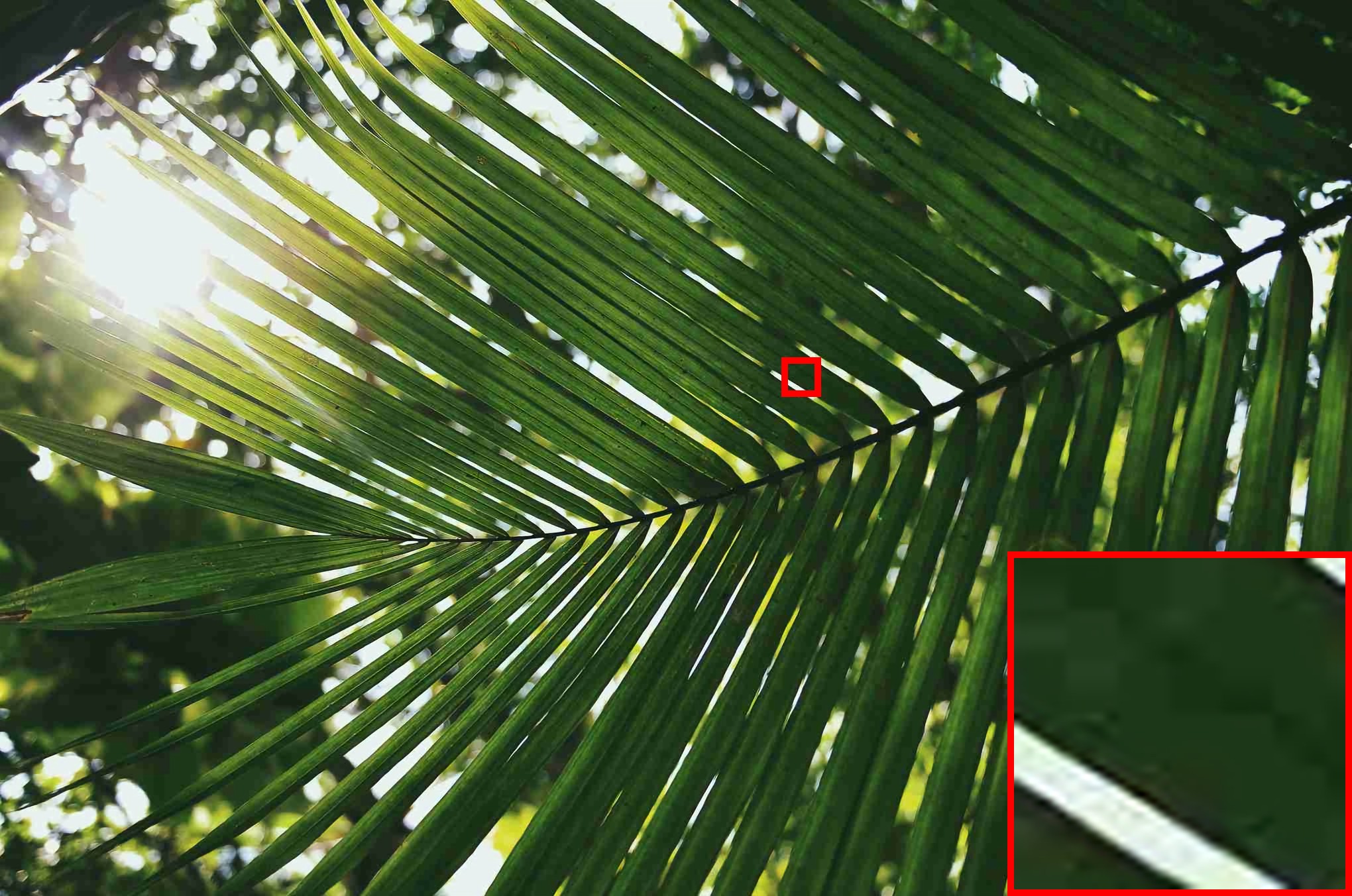} \label{subfig:deblock2} }%
   
     \caption{Qualitative evaluation of the effect of the knowledge sharing in our continual learning framework. 'Image 1' represents the outputs of the model without any knowledge sharing. 'Image 2' depicts the results of those models where the respective tasks are trained at last using the knowledge of all the previous tasks. (Zoom for the best view.)}
     \vspace{-3mm}
     \label{fig: all_image}
 \end{figure*}

\subsubsection{Subjective Comparison}
Figure~\ref{fig: all_image} shows the subjective comparison of restoration performance. We use the model which allocates $1.25\%$ of trainable parameters for each restoration operation. We present the qualitative analysis of a method that does not utilize knowledge from previous tasks, alongside a method that leverages knowledge from past restoration tasks. The results of the method that does not share the knowledge are termed 'Image-1' and the results of the knowledge sharing are termed 'Image-2', as shown in Figure~\ref{fig: all_image}. In the case of knowledge sharing, we use the trained model of a task that has been trained at the end. Therefore, we can observe significant visual differences. In the case of blurring and deblocking, the red boxes highlight the performance improvement in both Figure~\ref{subfig:deblock2} and Figure~\ref{subfig:deblur2} as compared to Figure~\ref{subfig:deblock1} and Figure~\ref{subfig:deblur1}, respectively. In the denoising task, Figure~\ref{subfig:denoise2} shows better results with fewer artifacts as compared to Figure~\ref{subfig:denoise1}. We also observe a lesser rain streak effect in Figure~\ref{subfig:derain2} as compared to Figure~\ref{subfig:derain1}. 

\subsubsection{Shuffling the Training order}

\begin{figure}[!htb]
    \centering
    \includegraphics[width=0.47\textwidth]{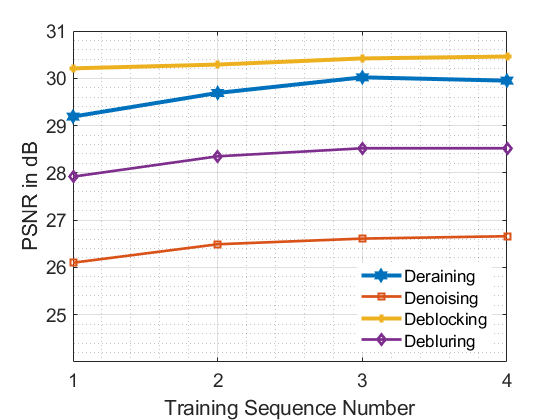}
    \caption{Graphical representation of the effect of training sequence on each task. As the training sequence number of a task is increased, the performance of the task increases due to better adoption of previous tasks' knowledge.}
    \label{fig: sequence}
\end{figure}

Previously, we demonstrated only one sequence of different restoration tasks, which are trained continually, and found that the performance of a restoration task improves if we share the knowledge from past tasks. Here, we shuffle the sequence of the four restoration tasks in such a manner that each task occupies every available position in the sequence and perform the experiments. Table~\ref{tab:continual_comp} shows the quantitative evaluation of four different experiments where each task is in a different training sequence position in each experiment. For example, deraining is the first task in Experiment 1, while it is the last task in Experiment 2, and it is the second and third tasks in Experiment 3 and Experiment 4, respectively. Figure~\ref{fig: sequence} depicts the graphical representation of Table~\ref{tab:continual_comp}. We observe that the performance of the task improves as the training sequence number increases. The training sequence number of a restoration task is the position at which it appears in training order while training the tasks continually. If a particular task's training sequence position is $4$, it means the model has already been trained on three different restoration tasks and utilizes the shared knowledge from those three previous tasks for the current task. We can conclude two things from these experiments. Firstly, the knowledge of previously learned tasks plays a crucial role in lifelong learning, and our method successfully helps future tasks to adopt the knowledge of the past tasks. Secondly, the training sequence plays a crucial role. The performance of the first task always degrades due to the non-availability of previous knowledge. Therefore, backward Continual Learning, in which past-trained tasks can fine-tune their knowledge from future tasks, will be able to mitigate the effect of the training sequence. We plan to explore it in future work.
\begin{table}[!htb]
\centering
\resizebox{0.48\textwidth}{!}{%
\begin{tabular}{|c|cc|cc|cc|cc|}
\hline
           & \multicolumn{2}{c|}{Experiment 1} & \multicolumn{2}{c|}{Experiment 2} & \multicolumn{2}{c|}{Experiment 3} & \multicolumn{2}{c|}{Experiment 4} \\ \hline
 &
  \multicolumn{1}{c|}{\begin{tabular}[c]{@{}c@{}}Training\\ Sequence\end{tabular}} &
  \begin{tabular}[c]{@{}c@{}}PSNR\\ in dB\end{tabular} &
  \multicolumn{1}{c|}{\begin{tabular}[c]{@{}c@{}}Training\\ Sequence\end{tabular}} &
  \begin{tabular}[c]{@{}c@{}}PSNR\\ in dB\end{tabular} &
  \multicolumn{1}{c|}{\begin{tabular}[c]{@{}c@{}}Training\\ Sequence\end{tabular}} &
  \begin{tabular}[c]{@{}c@{}}PSNR\\ in dB\end{tabular} &
  \multicolumn{1}{c|}{\begin{tabular}[c]{@{}c@{}}Training\\ Sequence\end{tabular}} &
  \begin{tabular}[c]{@{}c@{}}PSNR\\ in dB\end{tabular} \\ \hline
Deraining  & \multicolumn{1}{c|}{1}   & 29.19  & \multicolumn{1}{c|}{4}   & 29.95  & \multicolumn{1}{c|}{3}   & 30.02  & \multicolumn{1}{c|}{2}   & 29.69  \\ \hline
Denoising  & \multicolumn{1}{c|}{2}   & 26.49  & \multicolumn{1}{c|}{1}   & 26.10  & \multicolumn{1}{c|}{4}   & 26.66  & \multicolumn{1}{c|}{3}   & 26.61  \\ \hline
Deblocking & \multicolumn{1}{c|}{3}   & 30.42  & \multicolumn{1}{c|}{2}   & 30.29  & \multicolumn{1}{c|}{1}   & 30.21  & \multicolumn{1}{c|}{4}   & 30.46  \\ \hline
Debluring  & \multicolumn{1}{c|}{4}   & 28.52  & \multicolumn{1}{c|}{3}   & 28.52  & \multicolumn{1}{c|}{2}   & 28.35  & \multicolumn{1}{c|}{1}   & 27.92  \\ \hline
\end{tabular}%
}
\caption{Quantitative evaluation of the effect of training sequence. In each experiment, the training sequences are shuffled.}
\vspace{-3mm}
\label{tab:continual_comp}
\end{table}

\begin{table}[!htb]
\centering
\resizebox{0.45\textwidth}{!}{%
\begin{tabular}{|c|c|c|c|c|c|}
\hline
\begin{tabular}[c]{@{}c@{}}Knowledge\\ Sharing\end{tabular} &
  \begin{tabular}[c]{@{}c@{}}Noise\\ 10\end{tabular} &
  \begin{tabular}[c]{@{}c@{}}Noise\\ 20\end{tabular} &
  \begin{tabular}[c]{@{}c@{}}Noise\\ 30\end{tabular} &
  \begin{tabular}[c]{@{}c@{}}Noise\\ 40\end{tabular} &
  \begin{tabular}[c]{@{}c@{}}Real\\ Noise\end{tabular} \\ \hline
\ding{55} &
  33.75 &
  30.44 &
  28.22 &
  27.12 &
  36.90 \\ \hline
\ding{51} &
  33.75 &
  \textbf{30.75} &
  \textbf{29.01} &
  \textbf{27.82} &
  \textbf{37.63} \\ \hline
\end{tabular}%
}
\caption{Performance of continual task adaptation for $5$ separate denoising tasks.}
\vspace{-3mm}
\label{tab: noise}
\end{table}

\subsubsection{Similar restoration tasks}

Till now, we have only considered completely different restoration tasks. To analyze the continual restoration task adaptation in similar kinds of tasks, we performed experiments sequentially on four Gaussian noise levels: $ 10$, $20$, $30$, and $40$, followed by real noise. For training on real-world noise, we use the popular SIDD dataset~\cite{SIDD}, and clean images from the DIV2K dataset~\cite{agustsson2017ntire} are used to generate Gaussian noisy images. Table~\ref{tab: noise} shows the performance of our CMC module-based continual learning framework in those tasks. The experimental setup is the same as the previous one, as performed in Table~\ref{tab:continual}. $20\%$ of model parameters are allocated for each task. We can observe from Table~\ref{tab: noise} that adopting knowledge from previous easy denoising tasks (i.e., less noisy images) significantly improves the performance on complex denoising tasks (i.e., heavily noisy images) and real noise. The experimental results further validate the feasibility of our proposed method. 
\begin{table}[!htb]
\centering
\resizebox{0.45\textwidth}{!}{%
\begin{tabular}{|c|c|c|c|c|}
\hline
\begin{tabular}[c]{@{}c@{}}Parameter\\ Expansion\end{tabular} & Derain & Denoise & Deblocking & Deblur \\ \hline
\ding{55}                                                            & 29.19     & 26.49      & 30.42         & 28.13     \\ \hline
\ding{51}                                                           & \textbf{29.51}     & \textbf{26.65}      & \textbf{30.51}         & \textbf{28.72}     \\ \hline
\end{tabular}%
}
\caption{Quantitative performance analysis when the parameter of some key layer of restoration is increased without hampering the computational performance.}
\label{tab:extension}
\vspace{-3mm}
\end{table}

\subsubsection{Parameter expansion}

In a deep learning network, some layers can demand more parameters to learn key essential features. Allocating fewer parameters in those layers may result in the loss of crucial features, ultimately leading to poor performance. In the case of restoration tasks, the input and output layers play a crucial role, as the first one extracts valuable features from the image, and the second one reconstructs the image. The number of kernel parameters is generally less in those layers as compared to other layers. This is because the input layer maps from $3-D$ RGB image and the output layer maps to $3-D$ RGB image. Therefore, if we allocate the same percentage of free parameters in those layers similar to other layers, it will lead to poor performance. However, allocating a higher percentage in those layers will lead to early parameter exhaustion in those layers. Using our method, we can easily expand the parameter space of the input and output layers without any computational overhead. Table~\ref{tab:extension} shows the performance of parameter expansion in those two layers. All models in that table use only $1.25\%$ of parameters. All the layers use the CMC-5 module, except for the first and final layers, which use the CMC-10 module. We can observe from the table that the performance increases significantly across all restoration tasks with the flexible and straightforward modifications that our module offers.
\begin{table}[!htb]
\centering
\resizebox{0.47\textwidth}{!}{%
\begin{tabular}{|c|c|c|c|c|c|}
\hline
Method &
  \begin{tabular}[c]{@{}c@{}}Trainable\\ Parameters\end{tabular} &
  \begin{tabular}[c]{@{}c@{}}Kernel\\ Parameters\end{tabular} &
  Memory &
  \begin{tabular}[c]{@{}c@{}}Flops\\ (GMac)\end{tabular} &
  \begin{tabular}[c]{@{}c@{}}Inference\\ Time\end{tabular} \\ \hline
\multirow{3}{*}{Type-1}                       & $1\times$ & $1\times$ & 2083 MB & 36.86 & 4.220 ms \\ 
                                              & $1.8\times$ & $1.8\times$ & 2099 MB & 65.4 & 9.080 ms \\ 
                                              & $4\times$ & $4\times$ & 2099 MB & 146.57 & 20.39 ms \\ \hline
\multicolumn{1}{|l|}{\multirow{3}{*}{Type-2}} & $1\times$ & $1\times$ & 2083 MB & 36.86 & 4.220 ms \\  
\multicolumn{1}{|l|}{}                        & $2\times$ & $2\times$ & 2571 MB & 73.73 & 8.49 ms \\  
\multicolumn{1}{|l|}{}                        & $4\times$ & $4\times$ & 3547 MB & 147.46 & 16.42 ms \\ \hline
\multirow{3}{*}{\begin{tabular}[c]{@{}c@{}}CMC-n\\ (Ours) \end{tabular}}                         & $1\times$ & $1\times$ & 2093 MB & 36.864 & 4.193 ms \\ 
                                              & $2\times$ & $1\times$ & 2111 MB & 36.87 & 4.237 ms \\ 
                                              & $4\times$ & $1\times$ & 2111 MB & 36.88 & 4.255 ms \\ \hline
\end{tabular}%
}
\caption{Computational complexity analysis of different fundamental continual learning mechanisms under the premises of parameter expansion}
\vspace{-3mm}
\label{tab:compute}
\end{table}

\subsubsection{Computational Complexity Analysis}

The primary advantage of this proposed method is its reduction of computational overhead for lifelong learning, as there are inherent limitations in hardware and computing power. In Table~\ref{tab:compute}, we compare the computational overhead of our proposed CMC module and different continual learning ideas. In this experiment, we consider a convolution layer that consists of the same kernel parameters. We consider that as a base. Now, we assume that all the parameters of the layer have already been occupied by different tasks. Therefore, we need to increase the number of parameters. We calculate and compare the computational overhead when increasing the number of parameters by a factor of $2\times$ and $4\times$. We take two fundamental ways to increase the parameters in the literature. The first one is termed Type-1 in Table~\ref{tab:compute}, where the kernel size is increased to accommodate a larger number of free parameters~\cite{hung2019compacting}. The second one is termed Type-2, where the number of kernels is increased or a new layer is introduced for adding new parameters~\cite {rusu2016progressive,liu2020lira}. In this experimental setup, we use a convolutional layer with a $3\times 3$ kernel size as the base layer. It has $64$ input and $64$ output channels. Now, this layer processes $64$ input features, which have spatial dimensions of $1000\times 1000$. In Type-1, we increase the kernel size to $4\times 4$ and $6\times 6$ from $3\times 3$, thereby increasing the parameters by $1.8\times$ and $4\times$, respectively. A new convolution layer is introduced to increase the parameters in Type-2. In our case, we use CMC-5 as a base layer and use CMC-10 and CMC-20 to double and quadruple the number of parameters. It can be clearly seen from Table~\ref{tab:compute} that the CMC module does not significantly burden the memory requirements, FLOPs, and inference time. However, Type-1 increases the inference time and Flops significantly as we increase the parameters. This is because it increases the kernel size, which exponentially increases the computational burden. Type-2 drastically increases both the memory required during processing and inference time. As the CMC module does not significantly increase the inference time and memory requirement, it can serve the purpose of lifelong training.

\begin{table}[!htb]
\centering
\resizebox{0.48\textwidth}{!}{%
\begin{tabular}{|c|c|c|c|c|c|}
\hline
Model & \begin{tabular}[c]{@{}c@{}}Knowledge\\ Sharing\end{tabular} & denoise & derain & deblock & deblur \\ \hline
\multirow{2}{*}{RDN}   & \ding{55}  & 29.43/ 0.902 & 26.36/ 0.731 & 30.20/ 0.861 & 27.37/ 0.782 \\ \cline{2-6} 
                       & \ding{51} & 29.43/ 0.902 & \textbf{26.47/ 0.734} & \textbf{30.38/ 0.866} & \textbf{27.90/ 0.799} \\ \hline
\multirow{2}{*}{Dense} & \ding{55}  & 28.84/ 0.889 & 25.98/ 0.709 & 30.16/ 0.860 & 27.90/ 0.799 \\ \cline{2-6} 
                       & \ding{51} & 28.84/ 0.889 & \textbf{26.39/ 0.728} & \textbf{30.36/ 0.865} & \textbf{28.60/ 0.815} \\ \hline
\end{tabular}%
}
\caption{Performance of continual task adaptation on two different model architectures.}
\vspace{-3mm}
\label{tab: model}
\end{table}

\subsubsection{Adopting CMC in different architectures}

To prove the extendibility of our proposed CMC in different deep architectures, we consider two popular network topologies for experiment purposes, namely dense block~\cite{dense, srdense} and residual dense network (RDN)~\cite{rdn}. Table~\ref{tab: model} shows the performance of our proposed continual learning framework on two different network architectures. The experimental setup is the same as the previous one, as performed in Table~\ref{tab:continual}. We consider $6$ blocks for both dense and RDN block-based architecture. We only consider $1.25\%$ of model parameters for each task. We can observe from Table~\ref{tab: model} that both architectures follow a similar trend, as we witness in the residual architecture. We provide both PSNR and SSIM values, and our experimental results show that SSIM follows a similar trend to PSNR.

\section{Limitations, Impact and Future Work}
\label{sec:future}
Our proposed Continual Memory in Convolution (CMC) module serves the purpose of lifelong learning, as it allows us to add knowledge without forgetting, and it does not impose an extra computational burden on the compact system. However, the main drawback of this approach is the number of parameters. More parameters are required in the CMC module to produce the same performance as compared to a conventional convolution layer. But nowadays, the system memory in a compact system is easily extendable. Therefore, our module can easily work in those systems. Currently, our model can only reuse the knowledge from past tasks. However, we believe that with a simple modification, this module has tremendous potential to learn backwards, i.e., to improve past restoration performance by utilizing knowledge from future tasks. Handling multiple known degradations in an image by leveraging knowledge of individual degradations can be explored through knowledge sharing by fusing the knowledge of individual tasks. These ideas can be the future scope of this work.

\section{Conclusion}
\label{sec:conclude}

In this paper, we propose a modification of the conventional convolution layer. By making this simple modification, we can continuously adapt the learned experience from previous tasks and share those experiences with the current task to improve its performance. We address the shortcomings of lifelong learning for image restoration tasks, and our module serves as a prospective solution. This is the first-of-a-kind work where diverse restoration tasks have been handled through continual learning. The proposed mechanism shares knowledge across tasks without changing the backbone architecture. The knowledge base can be continuously expanded with a minimal computational burden. We experimentally observe the benefits of knowledge sharing between completely different restoration tasks, as it helps to improve the performance by a significant margin. 

%% file: mainbib.bib
@String(CVPR= {IEEE Conf. Comput. Vis. Pattern Recog.})

@String(ECCV= {Eur. Conf. Comput. Vis.})

@String(ICIP = {IEEE Int. Conf. Image Process.})

@String(ICLR = {Int. Conf. Learn. Represent.})

@String(CVPRW= {IEEE Conf. Comput. Vis. Pattern Recog. Worksh.})

@String(CVPR  = {CVPR})

@String(ECCV  = {ECCV})

@String(ICIP  = {ICIP})

@String(ICLR  = {ICLR})

@String(CVPRW= {CVPRW})

@inproceedings{dense,
  title={Densely connected convolutional networks},
  author={Huang, Gao and Liu, Zhuang and Van Der Maaten, Laurens and Weinberger, Kilian Q},
  booktitle={Proceedings of the IEEE conference on computer vision and pattern recognition},
  pages={4700--4708},
  year={2017}
}

@inproceedings{srdense,
  title={Image super-resolution using dense skip connections},
  author={Tong, Tong and Li, Gen and Liu, Xiejie and Gao, Qinquan},
  booktitle={Proceedings of the IEEE international conference on computer vision},
  pages={4799--4807},
  year={2017}
}

@article{rdn,
  title={Residual dense network for image restoration},
  author={Zhang, Yulun and Tian, Yapeng and Kong, Yu and Zhong, Bineng and Fu, Yun},
  journal={IEEE Transactions on Pattern Analysis and Machine Intelligence},
  volume={43},
  number={7},
  pages={2480--2495},
  year={2020},
  publisher={IEEE}
}

@INPROCEEDINGS{SIDD,
  author={Abdelhamed, Abdelrahman and Lin, Stephen and Brown, Michael S.},
  booktitle={2018 IEEE/CVF Conference on Computer Vision and Pattern Recognition}, 
  title={A High-Quality Denoising Dataset for Smartphone Cameras}, 
  year={2018},
  volume={},
  number={},
  pages={1692-1700},
  doi={10.1109/CVPR.2018.00182}}

@InProceedings{MAS,
author = {Aljundi, Rahaf and Babiloni, Francesca and Elhoseiny, Mohamed and Rohrbach, Marcus and Tuytelaars, Tinne},
title = {Memory Aware Synapses: Learning what (not) to forget },
booktitle = {The European Conference on Computer Vision (ECCV)},
month = {September},
year = {2018}
}

@inproceedings{liu2020lira,
  title={LIRA: Lifelong Image Restoration from Unknown Blended Distortions},
  author={Liu, Jianzhao and Lin, Jianxin and Li, Xin and Zhou, Wei and Liu, Sen and Chen, Zhibo},
  booktitle={Computer Vision--ECCV 2020: 16th European Conference, Glasgow, UK, August 23--28, 2020, Proceedings, Part XVIII 16},
  pages={616--632},
  year={2020},
  organization={Springer}
}

@InProceedings{Zhou_2021_CVPR,
    author    = {Zhou, Man and Xiao, Jie and Chang, Yifan and Fu, Xueyang and Liu, Aiping and Pan, Jinshan and Zha, Zheng-Jun},
    title     = {Image De-Raining via Continual Learning},
    booktitle = {Proceedings of the IEEE/CVF Conference on Computer Vision and Pattern Recognition (CVPR)},
    month     = {June},
    year      = {2021},
    pages     = {4907-4916}
}

@InProceedings{Fu_2017_CVPR,
author = {Fu, Xueyang and Huang, Jiabin and Zeng, Delu and Huang, Yue and Ding, Xinghao and Paisley, John},
title = {Removing Rain From Single Images via a Deep Detail Network},
booktitle = {Proceedings of the IEEE Conference on Computer Vision and Pattern Recognition (CVPR)},
month = {July},
year = {2017}
}

@InProceedings{Gandelsman_2019_CVPR,
author = {Gandelsman, Yosef and Shocher, Assaf and Irani, Michal},
title = {"Double-DIP": Unsupervised Image Decomposition via Coupled Deep-Image-Priors},
booktitle = {Proceedings of the IEEE/CVF Conference on Computer Vision and Pattern Recognition (CVPR)},
month = {June},
year = {2019}
}

@article{li2020zero,
  title={Zero-shot image dehazing},
  author={Li, Boyun and Gou, Yuanbiao and Liu, Jerry Zitao and Zhu, Hongyuan and Zhou, Joey Tianyi and Peng, Xi},
  journal={IEEE Transactions on Image Processing},
  volume={29},
  pages={8457--8466},
  year={2020},
  publisher={IEEE}
}

@inproceedings{chang2020spatial,
  title={Spatial-adaptive network for single image denoising},
  author={Chang, Meng and Li, Qi and Feng, Huajun and Xu, Zhihai},
  booktitle={European Conference on Computer Vision},
  pages={171--187},
  year={2020},
  organization={Springer}
}

@inproceedings{liu2019dual,
  title={Dual residual networks leveraging the potential of paired operations for image restoration},
  author={Liu, Xing and Suganuma, Masanori and Sun, Zhun and Okatani, Takayuki},
  booktitle={Proceedings of the IEEE/CVF Conference on Computer Vision and Pattern Recognition},
  pages={7007--7016},
  year={2019}
}

@inproceedings{zamir2021multi,
  title={Multi-stage progressive image restoration},
  author={Zamir, Syed Waqas and Arora, Aditya and Khan, Salman and Hayat, Munawar and Khan, Fahad Shahbaz and Yang, Ming-Hsuan and Shao, Ling},
  booktitle={Proceedings of the IEEE/CVF Conference on Computer Vision and Pattern Recognition},
  pages={14821--14831},
  year={2021}
}

@inproceedings{chen2021hinet,
  title={HINet: Half instance normalization network for image restoration},
  author={Chen, Liangyu and Lu, Xin and Zhang, Jie and Chu, Xiaojie and Chen, Chengpeng},
  booktitle={Proceedings of the IEEE/CVF Conference on Computer Vision and Pattern Recognition},
  pages={182--192},
  year={2021}
}

@inproceedings{kar2021zero,
  title={Zero-Shot Single Image Restoration Through Controlled Perturbation of Koschmieder's Model},
  author={Kar, Aupendu and Dhara, Sobhan Kanti and Sen, Debashis and Biswas, Prabir Kumar},
  booktitle={Proceedings of the IEEE/CVF Conference on Computer Vision and Pattern Recognition},
  pages={16205--16215},
  year={2021}
}

@article{kirkpatrick2017overcoming,
  title={Overcoming catastrophic forgetting in neural networks},
  author={Kirkpatrick, James and Pascanu, Razvan and Rabinowitz, Neil and Veness, Joel and Desjardins, Guillaume and Rusu, Andrei A and Milan, Kieran and Quan, John and Ramalho, Tiago and Grabska-Barwinska, Agnieszka and others},
  journal={Proceedings of the national academy of sciences},
  volume={114},
  number={13},
  pages={3521--3526},
  year={2017},
  publisher={National Acad Sciences}
}

@article{li2017learning,
  title={Learning without forgetting},
  author={Li, Zhizhong and Hoiem, Derek},
  journal={IEEE transactions on pattern analysis and machine intelligence},
  volume={40},
  number={12},
  pages={2935--2947},
  year={2017},
  publisher={IEEE}
}

@inproceedings{dhar2019learning,
  title={Learning without memorizing},
  author={Dhar, Prithviraj and Singh, Rajat Vikram and Peng, Kuan-Chuan and Wu, Ziyan and Chellappa, Rama},
  booktitle={Proceedings of the IEEE/CVF Conference on Computer Vision and Pattern Recognition},
  pages={5138--5146},
  year={2019}
}

@inproceedings{aljundi2018memory,
  title={Memory aware synapses: Learning what (not) to forget},
  author={Aljundi, Rahaf and Babiloni, Francesca and Elhoseiny, Mohamed and Rohrbach, Marcus and Tuytelaars, Tinne},
  booktitle={Proceedings of the European Conference on Computer Vision (ECCV)},
  pages={139--154},
  year={2018}
}

@inproceedings{mallya2018packnet,
  title={Packnet: Adding multiple tasks to a single network by iterative pruning},
  author={Mallya, Arun and Lazebnik, Svetlana},
  booktitle={Proceedings of the IEEE conference on Computer Vision and Pattern Recognition},
  pages={7765--7773},
  year={2018}
}

@inproceedings{rebuffi2017icarl,
  title={icarl: Incremental classifier and representation learning},
  author={Rebuffi, Sylvestre-Alvise and Kolesnikov, Alexander and Sperl, Georg and Lampert, Christoph H},
  booktitle={Proceedings of the IEEE conference on Computer Vision and Pattern Recognition},
  pages={2001--2010},
  year={2017}
}

@article{shin2017continual,
  title={Continual learning with deep generative replay},
  author={Shin, Hanul and Lee, Jung Kwon and Kim, Jaehong and Kim, Jiwon},
  journal={arXiv preprint arXiv:1705.08690},
  year={2017}
}

@inproceedings{brahma2018subset,
  title={Subset replay based continual learning for scalable improvement of autonomous systems},
  author={Brahma, Pratik Prabhanjan and Othon, Adrienne},
  booktitle={2018 IEEE/CVF Conference on Computer Vision and Pattern Recognition Workshops (CVPRW)},
  pages={1179--11798},
  year={2018},
  organization={IEEE}
}

@inproceedings{hu2018overcoming,
  title={Overcoming catastrophic forgetting for continual learning via model adaptation},
  author={Hu, Wenpeng and Lin, Zhou and Liu, Bing and Tao, Chongyang and Tao, Zhengwei and Ma, Jinwen and Zhao, Dongyan and Yan, Rui},
  booktitle={International Conference on Learning Representations},
  year={2018}
}

@inproceedings{ostapenko2019learning,
  title={Learning to remember: A synaptic plasticity driven framework for continual learning},
  author={Ostapenko, Oleksiy and Puscas, Mihai and Klein, Tassilo and Jahnichen, Patrick and Nabi, Moin},
  booktitle={Proceedings of the IEEE/CVF Conference on Computer Vision and Pattern Recognition},
  pages={11321--11329},
  year={2019}
}

@article{lopez2017gradient,
  title={Gradient episodic memory for continual learning},
  author={Lopez-Paz, David and Ranzato, Marc'Aurelio},
  journal={Advances in neural information processing systems},
  volume={30},
  pages={6467--6476},
  year={2017}
}

@article{wu2018memory,
  title={Memory replay gans: Learning to generate new categories without forgetting},
  author={Wu, Chenshen and Herranz, Luis and Liu, Xialei and van de Weijer, Joost and Raducanu, Bogdan and others},
  journal={Advances in Neural Information Processing Systems},
  volume={31},
  pages={5962--5972},
  year={2018}
}

@article{wu2018incremental,
  title={Incremental classifier learning with generative adversarial networks},
  author={Wu, Yue and Chen, Yinpeng and Wang, Lijuan and Ye, Yuancheng and Liu, Zicheng and Guo, Yandong and Zhang, Zhengyou and Fu, Yun},
  journal={arXiv preprint arXiv:1802.00853},
  year={2018}
}

@inproceedings{titsias2020functional,
  title={Functional Regularisation for Continual Learning with Gaussian Processes},
  author={Titsias, Michalis K and Schwarz, Jonathan and Matthews, Alexander G de G and Pascanu, Razvan and Teh, Yee Whye},
  booktitle={ICLR},
  year={2020}
}

@article{rusu2016progressive,
  title={Progressive neural networks},
  author={Rusu, Andrei A and Rabinowitz, Neil C and Desjardins, Guillaume and Soyer, Hubert and Kirkpatrick, James and Kavukcuoglu, Koray and Pascanu, Razvan and Hadsell, Raia},
  journal={arXiv preprint arXiv:1606.04671},
  year={2016}
}

@article{rosenfeld2018incremental,
  title={Incremental learning through deep adaptation},
  author={Rosenfeld, Amir and Tsotsos, John K},
  journal={IEEE transactions on pattern analysis and machine intelligence},
  volume={42},
  number={3},
  pages={651--663},
  year={2018},
  publisher={IEEE}
}

@article{hung2019compacting,
  title={Compacting, picking and growing for unforgetting continual learning},
  author={Hung, Steven CY and Tu, Cheng-Hao and Wu, Cheng-En and Chen, Chien-Hung and Chan, Yi-Ming and Chen, Chu-Song},
  journal={arXiv preprint arXiv:1910.06562},
  year={2019}
}

@inproceedings{li2019learn,
  title={Learn to grow: A continual structure learning framework for overcoming catastrophic forgetting},
  author={Li, Xilai and Zhou, Yingbo and Wu, Tianfu and Socher, Richard and Xiong, Caiming},
  booktitle={International Conference on Machine Learning},
  pages={3925--3934},
  year={2019},
  organization={PMLR}
}

@inproceedings{xu2018reinforced,
  title={Reinforced Continual Learning},
  author={Xu, Ju and Zhu, Zhanxing},
  booktitle={NeurIPS},
  year={2018}
}

@inproceedings{gondara2016medical,
  title={Medical image denoising using convolutional denoising autoencoders},
  author={Gondara, Lovedeep},
  booktitle={2016 IEEE 16th international conference on data mining workshops (ICDMW)},
  pages={241--246},
  year={2016},
  organization={IEEE}
}

@inproceedings{svoboda2016cnn,
  title={CNN for license plate motion deblurring},
  author={Svoboda, Pavel and Hradi{\v{s}}, Michal and Mar{\v{s}}{\'\i}k, Luk{\'a}{\v{s}} and Zemc{\'\i}k, Pavel},
  booktitle={2016 IEEE International Conference on Image Processing (ICIP)},
  pages={3832--3836},
  year={2016},
  organization={IEEE}
}

@inproceedings{deng2020detail,
  title={Detail-recovery image deraining via context aggregation networks},
  author={Deng, Sen and Wei, Mingqiang and Wang, Jun and Feng, Yidan and Liang, Luming and Xie, Haoran and Wang, Fu Lee and Wang, Meng},
  booktitle={Proceedings of the IEEE/CVF conference on computer vision and pattern recognition},
  pages={14560--14569},
  year={2020}
}

@inproceedings{qu2019enhanced,
  title={Enhanced pix2pix dehazing network},
  author={Qu, Yanyun and Chen, Yizi and Huang, Jingying and Xie, Yuan},
  booktitle={Proceedings of the IEEE/CVF Conference on Computer Vision and Pattern Recognition},
  pages={8160--8168},
  year={2019}
}

@inproceedings{dong2020multi,
  title={Multi-scale boosted dehazing network with dense feature fusion},
  author={Dong, Hang and Pan, Jinshan and Xiang, Lei and Hu, Zhe and Zhang, Xinyi and Wang, Fei and Yang, Ming-Hsuan},
  booktitle={Proceedings of the IEEE/CVF Conference on Computer Vision and Pattern Recognition},
  pages={2157--2167},
  year={2020}
}

@inproceedings{tao2018scale,
  title={Scale-recurrent network for deep image deblurring},
  author={Tao, Xin and Gao, Hongyun and Shen, Xiaoyong and Wang, Jue and Jia, Jiaya},
  booktitle={Proceedings of the IEEE Conference on Computer Vision and Pattern Recognition},
  pages={8174--8182},
  year={2018}
}

@incollection{mccloskey1989catastrophic,
  title={Catastrophic interference in connectionist networks: The sequential learning problem},
  author={McCloskey, Michael and Cohen, Neal J},
  booktitle={Psychology of learning and motivation},
  volume={24},
  pages={109--165},
  year={1989},
  publisher={Elsevier}
}

@inproceedings{yang2017deep,
  title={Deep joint rain detection and removal from a single image},
  author={Yang, Wenhan and Tan, Robby T and Feng, Jiashi and Liu, Jiaying and Guo, Zongming and Yan, Shuicheng},
  booktitle={Proceedings of the IEEE conference on computer vision and pattern recognition},
  pages={1357--1366},
  year={2017}
}

@inproceedings{agustsson2017ntire,
  title={Ntire 2017 challenge on single image super-resolution: Dataset and study},
  author={Agustsson, Eirikur and Timofte, Radu},
  booktitle={Proceedings of the IEEE conference on computer vision and pattern recognition workshops},
  pages={126--135},
  year={2017}
}

@article{roth2009fields,
  title={Fields of experts},
  author={Roth, Stefan and Black, Michael J},
  journal={International Journal of Computer Vision},
  volume={82},
  number={2},
  pages={205},
  year={2009},
  publisher={Springer}
}

@inproceedings{he2016deep,
  title={Deep residual learning for image recognition},
  author={He, Kaiming and Zhang, Xiangyu and Ren, Shaoqing and Sun, Jian},
  booktitle={Proceedings of the IEEE conference on computer vision and pattern recognition},
  pages={770--778},
  year={2016}
}

@article{kingma2014adam,
  title={Adam: A method for stochastic optimization},
  author={Kingma, Diederik P and Ba, Jimmy},
  journal={arXiv preprint arXiv:1412.6980},
  year={2014}
}
